\documentclass[lettersize,journal]{IEEEtran}
\usepackage{amsmath,amsfonts}
\usepackage{algorithmic}
\usepackage{algorithm}
\usepackage{array}
\usepackage{textcomp}
\usepackage{stfloats}
\usepackage{url}
\usepackage{verbatim}
\usepackage{graphicx}
\usepackage{cite}
\usepackage{mathtools}
\usepackage[caption=false, font=footnotesize]{subfig}

\usepackage{multirow}
\usepackage[dvipsnames]{color,xcolor}
\usepackage{booktabs}
\usepackage{bbding}

\usepackage{pifont}

\begin{document}

\title{DualMamba: A Lightweight Spectral-Spatial Mamba-Convolution Network for Hyperspectral Image Classification
}

\author{Jiamu Sheng, Jingyi Zhou, Jiong Wang, Peng Ye, Jiayuan Fan,~\IEEEmembership{Member,~IEEE}\thanks{Jiamu Sheng and Jiayuan Fan are with the Academy for Engineering and Technology, Fudan University, Shanghai 200433, China (e-mail: jmsheng22@m.fudan.edu.cn; jyfan@fudan.edu.cn). \textit{(Corresponding author: Jiayuan Fan.)}}\thanks{Jingyi Zhou, Jiong Wang and Peng Ye are with School of Information Science and Technology, Fudan University, Shanghai 200433, China.}
} 


\maketitle

\maketitle


\begin{abstract}

The effectiveness and efficiency of modeling complex spectral-spatial relations are both crucial for Hyperspectral image (HSI) classification. Most existing methods based on CNNs and transformers still suffer from heavy computational burdens and have room for improvement in capturing the global-local spectral-spatial feature representation. To this end, we propose a novel lightweight parallel design called lightweight dual-stream Mamba-convolution network (DualMamba) for HSI classification. Specifically, a parallel lightweight Mamba and CNN block are first developed to extract global and local spectral-spatial features. First, the cross-attention spectral-spatial Mamba module is proposed to leverage the global modeling of Mamba at linear complexity. Within this module, dynamic positional embedding is designed to enhance the spatial location information of visual sequences. The lightweight spectral/spatial Mamba blocks comprise an efficient scanning strategy and a lightweight Mamba design to efficiently extract global spectral-spatial features. And the cross-attention spectral-spatial fusion is designed to learn cross-correlation and fuse spectral-spatial features. Second, the lightweight spectral-spatial residual convolution module is proposed with lightweight spectral and spatial branches to extract local spectral-spatial features through residual learning. Finally, the adaptive global-local fusion is proposed to dynamically combine global Mamba features and local convolution features for a global-local spectral-spatial representation. Compared with state-of-the-art HSI classification methods, experimental results demonstrate that DualMamba achieves significant classification accuracy on three public HSI datasets and a superior reduction in model parameters and floating point operations (FLOPs).


\end{abstract}


\section{Introduction}
\IEEEPARstart{H}{YPERSPECTRAL} image (HSI) classification is a key task in the field of remote sensing that involves categorizing each pixel in an image based on its spectral signature \cite{zhang2022artificial,mtmsd,a2snas,u2convformer}. The significance of HSI classification lies in its ability to utilize the rich information captured across hundreds of spectral bands to accurately identify materials and objects on the land. It has a wide range of applications in environmental monitoring \cite{obermeier2019grassland}, resource management \cite{wu2020resourse}, agriculture disaster response \cite{hsieh2020agriculture}, military defense \cite{shimoni2019military}, etc. 
\begin{figure}[t]
    \centering
    \subfloat[FLOPs]{\includegraphics[width = 0.5\columnwidth]{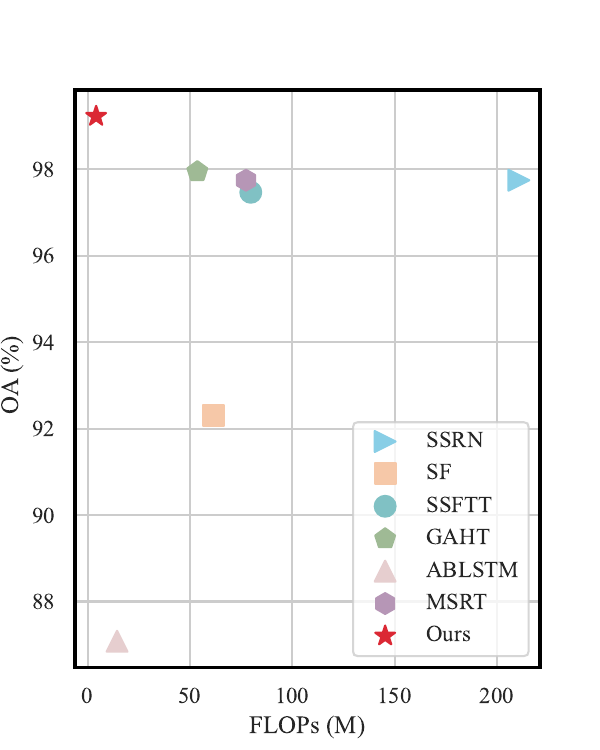}}
    \subfloat[Parameters]{\includegraphics[width = 0.5\columnwidth]{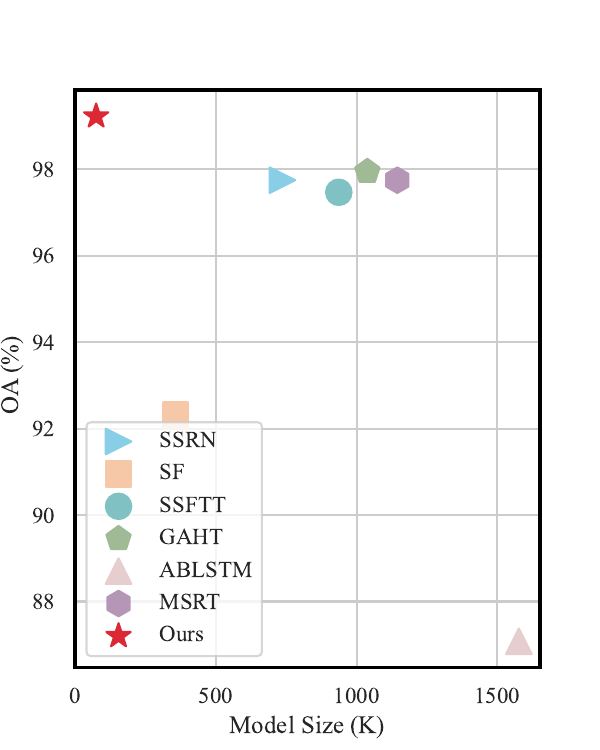}}
    \caption{Performance comparisons with respect to FLOPs and parameters on the Indian Pines dataset. Our DualMamba can outperform state-of-the-art methods with the fewest parameters and FLOPs.}
    \label{fig1}
    \vspace{-1em}
\end{figure}

With the advancement of deep learning, large amounts of HSI classification methods based on baseline models such as convolution neural networks (CNNs) \cite{yang20182dcnn,zhong2018ssrn,yu2021feedback,zhumingzhen} and transformers \cite{hong2021spectralformer,he2021vitspespa,segmsrt} have been extensively researched.
CNN-based methods leverage the capability of convolutions to extract local features, which is vital for recognizing textural patterns of objects in HSIs. Transformer-based methods benefit from the capacity of self-attention mechanisms to model long-range dependencies, effectively capturing global contextual information. Further, hybrid transformer-CNN architectures \cite{sun2022ssftt, mei2022gaht, wuke2023hyperspectral,zou2022lessformer, xu2023multiscale} have been studied to leverage the global modeling of transformer and local extraction of CNN, modeling spectral-spatial relations from global and local perspectives. Although the above deep learning-based methods have achieved promising results, they still suffer from crucial issues as follows.

First, existing HSI classification methods struggle to be both effective and efficient in global-local spectral-spatial modeling. CNN-based or transformer-based approaches can construct hierarchical structures by stacking CNN blocks or transformer encoders to capture local and global information from shallow and deep layers, respectively. However, this results in a large increase in model parameters, leading to inefficiency. Another approach to achieving global-local spectral-spatial modeling involves integrating global feature extractors, such as transformers, with local feature extractors, such as CNNs. Typically, existing methods employ a cascading manner to sequentially learn global/local relations \cite{sun2022ssftt,xu2023multiscale}. However, cascading structures tend to overlook global context during local modeling (and vice versa) and fail to decouple the discrimination of global and local features in modeling spectral-spatial relations \cite{xu2021vitae}. Therefore, we adopt a "divide and conquer" philosophy, considering a parallel structure to capture global and local spectral-spatial features, respectively.

Second, in global-local spectral-spatial modeling, utilizing transformers to capture global contextual information is inefficient. Self-attention mechanism within transformers requires computing attention N times for each token in a sequence of length N, resulting in a computational complexity of $\mathcal{O}(N^2)$ and an 
$N \times N$ attention matrix to store, which makes this approach inefficient in both time and space. Recently introduced as an alternative to transformers, the selective structured state space model (Mamba) \cite{gu2023mamba} has received significant attention in language tasks due to its global contextual modeling ability and efficiency, subquadratic-time computation, and linear memory complexity. Further, Mamba has been adapted for 2D vision tasks \cite{liu2024vmamba,zhu2024visionmamba,chenhao2024omni}, enhancing its global modeling capabilities for spatial contexts through a multi-directional scanning strategy. Naturally, we seek to employ Mamba in HSI classification as an alternative to transformers for efficiently modeling global spectral-spatial relations. However, directly applying existing vision Mamba methods to HSI classification does not yield satisfactory performance or computational efficiency due to several limitations:
\begin{enumerate}
\item These methods capture global context at the spatial level, inadequately capturing the rich spectral information in HSIs. Moreover, directly scanning features along the spectral dimension to extract global spectral features from Mamba introduces a heavy computational burden.
\item The multi-directional scanning strategy is redundant and inefficient for HSIs, leading to a large increase in model complexity and an exponential rise in both the number of parameters and FLOPs. Furthermore, since HSIs are captured by high-altitude sensors with a nadir viewing angle, spatial features obtained from scans at any angle are similar and redundant. 
\item The current design of the Vision Mamba block, which incorporates gated MLP and H3 blocks, introduces an excessive number of linear layers, thereby bringing a heavy computational burden. 
\end{enumerate}
Therefore, the structural design of the Mamba block should adequately consider both the effectiveness of global spectral-spatial feature extraction and lightweight efficiency.


Third, in local spectral-spatial modeling, existing CNN-based methods typically employ traditional 2D or 3D convolutions to extract spectral-spatial features \cite{zhong2018ssrn}. Since 2D and 3D convolutions process all features across both spatial and spectral dimensions simultaneously, they lead to a substantial increase in parameters and computational demands when dealing with a large number of spectral bands in HSIs. While some lightweight convolution designs increasingly reduce model parameters and complexity, such as depthwise convolutions and pointwise convolutions, they respectively extract only spatial or spectral features and fail to effectively model complex spectral-spatial relations. Therefore, designing both an effective and lightweight network that can extract local spectral-spatial features is crucial.

In view of these, we propose an efficient, lightweight dual-stream Mamba-convolution network for HSI classification, named DualMamba. Our key design is parallel lightweight Mamba and CNN to efficiently model complex spectral-spatial relations from both global and local perspectives. To extract global spectral-spatial features, we propose the cross-attention spectral-spatial Mamba module. This module initially incorporates dynamic positional embedding to enhance the spatial location information of visual sequences. Then, the embeddings are fed into our designed lightweight spatial/spectral Mamba blocks, respectively, extracting global spatial and spectral features. In the design of the lightweight spatial Mamba block, a spatial unidirectional scan is proposed to efficiently capture non-redundant global spatial features. Conversely, the lightweight spectral Mamba block is designed with a spectral bidirectional scan to learn global spectral correlations from both forward and reversed spectral sequences.
Subsequently, a cross-attention mechanism is employed to learn cross-correlations and fuse global spectral-spatial features. To extract local spectral-spatial features, we propose the lightweight spectral-spatial residual convolution module. This module utilizes a lightweight 3D convolution kernel and a depthwise convolution kernel to extract local spectral and spatial features, respectively, efficiently capturing local spectral-spatial relations through residual learning. After the dual-stream Mamba-convolution processing, we perform the adaptive global-local fusion that dynamically modulates the balance between global and local features in response to the specific contents of HSI, facilitating the learning of global-local spectral-spatial representation. Evaluated on three public datasets, our method outperforms state-of-the-art methods with minimal parameters and FLOPs, as illustrated in Fig~\ref{fig1}.

To summarize, our contributions are listed as follows.
\begin{itemize}
    \item[1)] We propose a novel lightweight dual-stream hybrid network that parallelly integrates Mamba for global context capture and a lightweight spectral-spatial CNN for local feature extraction, facilitating efficient and effective modeling of global-local spectral-spatial relations in HSI classification. Our method significantly outperforms existing state-of-the-art methods with minimal parameters and computational complexity.
    \item[2)] We propose the cross-attention spectral-spatial Mamba module. This module combines dynamic positional embeddings with parallel lightweight spectral/spatial Mamba blocks, employing a cross-attention fusion to efficiently model global spectral-spatial relations.
    \item[3)] We propose the lightweight spectral-spatial residual convolution module, which is designed with lightweight spectral and spatial branches to efficiently extract local spectral-spatial features through residual learning. 
    \item[4)] We propose the adaptive global-local fusion that dynamically adjusts the weighting between global and local spectral-spatial modeling based on HSI contents, learning a global-local spectral-spatial representation.
\end{itemize}

The remainder of this paper is organized as follows: Section II describes related work. Section III introduces our DualMamba in detail. Section IV conducts extensive experiments on three HSI datasets to demonstrate the effectiveness of the proposed method. Finally, Section V draws some conclusions.

\section{Related Work}
\subsection{Deep Learning-based Methods for HSI Classification}
Due to the significant advancements in deep learning for various computer vision tasks, an array of progressive deep learning-based networks has been widely adopted for HSI classification methods. CNNs particularly gain prominence by leveraging their capability to extract spatially structural and locally contextual information, thus becoming mainstream in HSI classification. Zhu \emph{et al.} \cite{zhong2018ssrn} incorporate a residual block within a sequential spectral-spatial feature learning network, enhanced by spectral and spatial attention modules. Zhu \emph{et al.} \cite{zhumingzhen} design a lightweight CNN network that utilizes multi-scale features for extracting contextual information.

RNNs, another fundamental neural network model, are extensively employed in HSI classification due to their proficiency in processing sequential data. These networks manage hyperspectral sequences and discern dependencies among spectral bands. Hang \emph{et al.} \cite{hang2019cascadedrnn} develop a cascaded RNN that treats hyperspectral pixels as sequential data, effectively capturing the relationships between adjacent spectral bands. Zhang \emph{et al.} \cite{zhang2018ssrnn} devise a method incorporating local spatial sequencing within an RNN framework to extract local and semantic information. Mei \emph{et al.} \cite{ablstm} propose a bidirectional Long Short-Term Memory (Bi-LSTM) network to investigate bidirectional spectral correlations in HSIs. Moreover, Zhou \emph{et al.} \cite{zhou2021multiscanningrnn} introduce a multiscanning strategy with RNNs, accentuating the sequential nature of HSI pixels and thoroughly addressing spatial dependencies within HSI patches.

Howerver, both CNNs and RNNs face challenges in modeling long-term spectral-spatial information dependencies. To overcome this limitation, researchers have explored the utility of transformers and have extensively applied them in HSI classification. Hong \emph{et al.} \cite{hong2021spectralformer} develop the SpectralFormer (SF), a transformer-based backbone network that efficiently processes the spectral sequence of neighboring bands, effectively capturing the sequence attributes of spectral signatures. To leverage both the strengths of transformers and CNN, cascading transformer-CNN architecture has emerged. Sun \emph{et al.} \cite{sun2022ssftt} introduce the SSFTT, which incorporates a Gaussian weighted token module into the transformer architecture to extract high-level semantic features, leading to improved performance. Zou \emph{et al.} \cite{zou2022lessformer} design the LESSFormer, including the HSI2Token module and a local-enhanced transformer encoder. This design adaptively generates spectral-spatial tokens and concurrently enhances local information while preserving global context. Additionally, Mei \emph{et al.} \cite{mei2022gaht} create the GAHT, which advances the extraction of local features in HSIs through a grouped pixel embedding module within the transformer framework, further optimizing feature representation.

Nevertheless, the aforementioned methods struggle to comprehensively model local-global spectral-spatial relations while maintaining a lightweight structure for computational efficiency. CNNs fail to simultaneously provide a large receptive field and a lightweight structure to extract local features. RNNs face challenges in modeling long-term global dependencies, and while transformers' core mechanism, self-attention, can effectively capture global contextual information, they typically require extensive resources due to their quadratic computational complexity. And cascading transformer-CNN methods ignore inherent relations between global and local modeling. Therefore, we explore the parallel Mamba-convolution architecture, which offers low time-memory complexity and exceptional global-local spectral-spatial feature extraction capabilities. 
\subsection{State Space Model and Mamba}
The State Space Model (SSM) \cite{gu2021s4,smith2022s5,gu2023mamba} is a fundamental model in control theory for modeling dynamic systems, enabling the representation of higher-order derivatives through only first-order derivatives and vector quantities. To address the high computational complexity of Transformers, Structured SSM (S4) \cite{gu2021s4} has been proposed to capture long dependencies within sequences based on the SSM framework. Although S4 operates with linear complexity, its parameters are static and independent of the input, which limits its ability to dynamically model effectively based on the input. Therefore, Mamba \cite{gu2023mamba} introduces an input-dependent parametrization approach for SSMs and includes a simple selection mechanism. Additionally, Mamba presents an efficient hardware-aware algorithm utilizing selective scanning, ensuring both effectiveness and efficiency in capturing global contextual information.

As Mamba succeeds in language tasks, Mamba-based vision models have gradually emerged. To adapt Mamba for processing image data, existing work has focused on devising scanning strategies that transform image data into sequential data. Zhu et al. \cite{zhu2024visionmamba} propose the Vision Mamba (ViM) architecture, which utilizes bidirectional scans to convert the image into an ordered visual sequence. Liu et al. \cite{liu2024vmamba} introduce VMamba, designing a cross-selective scan that scans images horizontally and vertically across the spatial dimension. Additionally, Zhao et al. \cite{chenhao2024omni} introduce an omnidirectional selective scan mechanism to capture large spatial features from various directions.
However, these models primarily focus on extracting global spatial context with Mamba and fail to effectively model the complex spectral-spatial relations in HSIs, which contain continuous spectral information. Furthermore, current multi-directional scanning strategies severely increase parameters and FLOPs, intensifying the computational burden and failing to meet the efficiency needs of HSI classification. Finally, HSI classification requires modeling spectral-spatial relations from both local and global perspectives. Therefore, designing Mamba-based methods for HSI classification necessitates a comprehensive consideration of these aspects.
\begin{figure*}
    \centering
    \includegraphics[width=\textwidth]{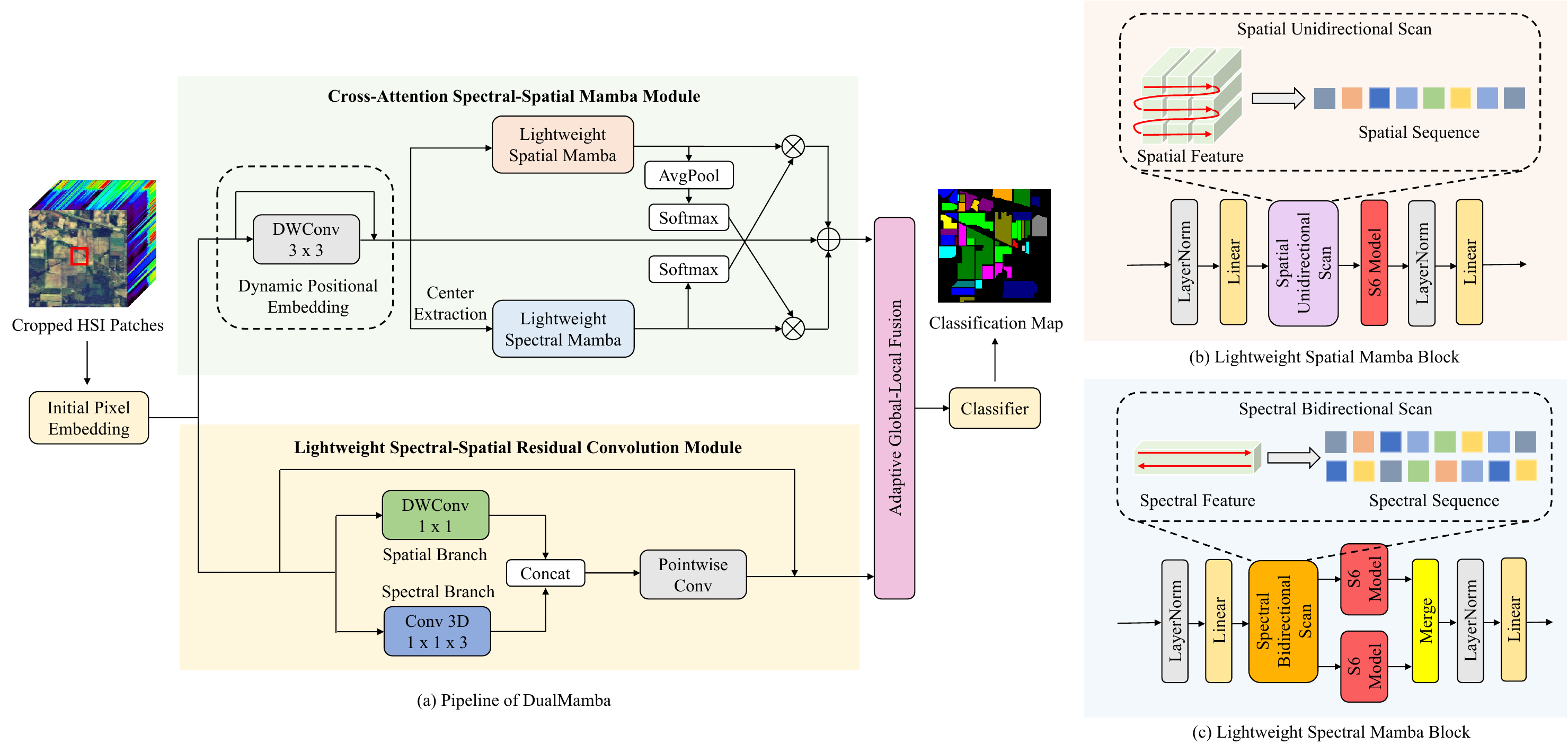}
    \caption{The overview of our proposed DualMamba in (a). The core of our method leverages the cross-attention spectral-spatial Mamba module and the lightweight spectral-spatial residual convolution module to efficiently extract spectral-spatial features from global and local perspectives, respectively. The structures of lightweight spatial Mamba and lightweight spatial Mamba are illustrated in (b) and (c), respectively. Subsequently, an adaptive global-local fusion module is employed to effectively achieve a comprehensive spectral-spatial representation.}
    \label{fig2}
\end{figure*}

\section{Method}
The overall architecture of our proposed DualMamba is depicted in Fig.~\ref{fig2}. 
In the following section, we introduce the proposed DualMamba in detail.

\subsection{Preliminary}
\subsubsection{State Space Model} 
The state space model (SSM) is a commonly used model in control theory, which utilizes an intermediate state variable to establish linear relations between other variables, the state variable, and inputs. Inspired by SSM, the Structured State Space Sequence Model (S4) is proposed for modeling sequential data. This model maps a one-dimensional sequence $x(t) \in \mathbb{R}$ to $ y(t) \in \mathbb{R}$ through a hidden state $h(t) \in \mathbb{R}^{N}$, which can be formulated by the following ordinary differential equation:
\begin{equation}
\begin{aligned}
h^{\prime}(t) &= \mathbf{A}h(t) + \mathbf{B}x(t),  \\
y(t) &= \mathbf{C}h(t),
\label{eq:ssm}
\end{aligned}
\end{equation}
where $\mathbf{A} \in \mathbb{R}^{N \times N}$ denotes the state transition matrix, $\mathbf{B} \in \mathbb{R}^{N \times 1}$ and $ \mathbf{C} \in \mathbb{R}^{1 \times N}$ respectively represent the matrices that map inputs to states and states to outputs.

Given that images and texts are discrete signals, the S4 model needs to be discretized using the Zero-Order Hold (ZOH) assumption. Specifically, the continuous-time parameters $\mathbf{A}, \mathbf{B}$ are transformed to their discretized counterparts $\overline{\mathbf{A}}, \overline{\mathbf{B}}$ with a timescale parameter $\Delta$ as follows:
\begin{equation}
\begin{aligned}
\overline{\mathbf{A}} &= \exp(\Delta\mathbf{A}), \\
\overline{\mathbf{B}} &= (\Delta\mathbf{A})^{-1}(\exp(\Delta\mathbf{A}) - \mathbf{I}) \cdot \Delta\mathbf{B}.
\end{aligned}
\end{equation}
Thus, the Eq.~\eqref{eq:ssm} can be reformulated as:
\begin{equation}
\begin{aligned}
h_t &= \overline{\mathbf{A}}h_{t-1} + \overline{\mathbf{B}}x_t, \\
y_t &= \mathbf{C}h_t.
\label{eq:ssm_d}
\end{aligned}
\end{equation}

To enhance the computational efficiency of the S4 model, the iterative process in Eq.~\eqref{eq:ssm_d} can be reformulated as a convolution: 
\begin{equation}
\begin{aligned}
\overline{\mathbf{K}} &= (\mathbf{C}\overline{\mathbf{B}}, \mathbf{C}\overline{\mathbf{AB}}, \ldots, \mathbf{C}\overline{\mathbf{A}}^{L-1}\overline{\mathbf{B}}), \\
\boldsymbol{y} &= \boldsymbol{x} * \overline{\mathbf{K}},
\label{eq:ssm_c}
\end{aligned}
\end{equation}
where $L$ denotes the length of the input sequence $\boldsymbol{x}$ and $\overline{\mathbf{K}} \in \mathbb{R}^{L}$ serves as the the S4 convolution kernel.
\subsubsection{Mamba}
The S4 model exhibits linear time complexity, yet it is constrained in representing sequence context due to time-invariant parameterization. Mamba introduces a selection mechanism and proposes the selective scan S4 (S6) model that redefines the parameterization to be input-dependent, thus enhancing dynamic interactions across sequences. Specifically, Mamba simply makes several parameters $\mathbf{B}, \mathbf{C}, \Delta$ functions of the input $\boldsymbol{x} \in \mathbb{R}^{B \times L \times D}$:
\begin{equation}
\mathbf{B}, \mathbf{C}, \Delta = Linear(\boldsymbol{x}),
\end{equation}
where $\mathbf{B} \in \mathbb{R}^{B \times L \times N}$, $\mathbf{C} \in \mathbb{R}^{B \times L \times N}$, and $\Delta \in \mathbb{R}^{B \times L \times D}$. Additionally, Mamba introduces a hardware-aware algorithm for efficient training. Our method utilizes the S6 model from Mamba, leveraging its computational efficiency and capability to effectively model global long-range dependencies.

\begin{figure}[t]
    \centering
    \includegraphics[width=0.85\columnwidth]{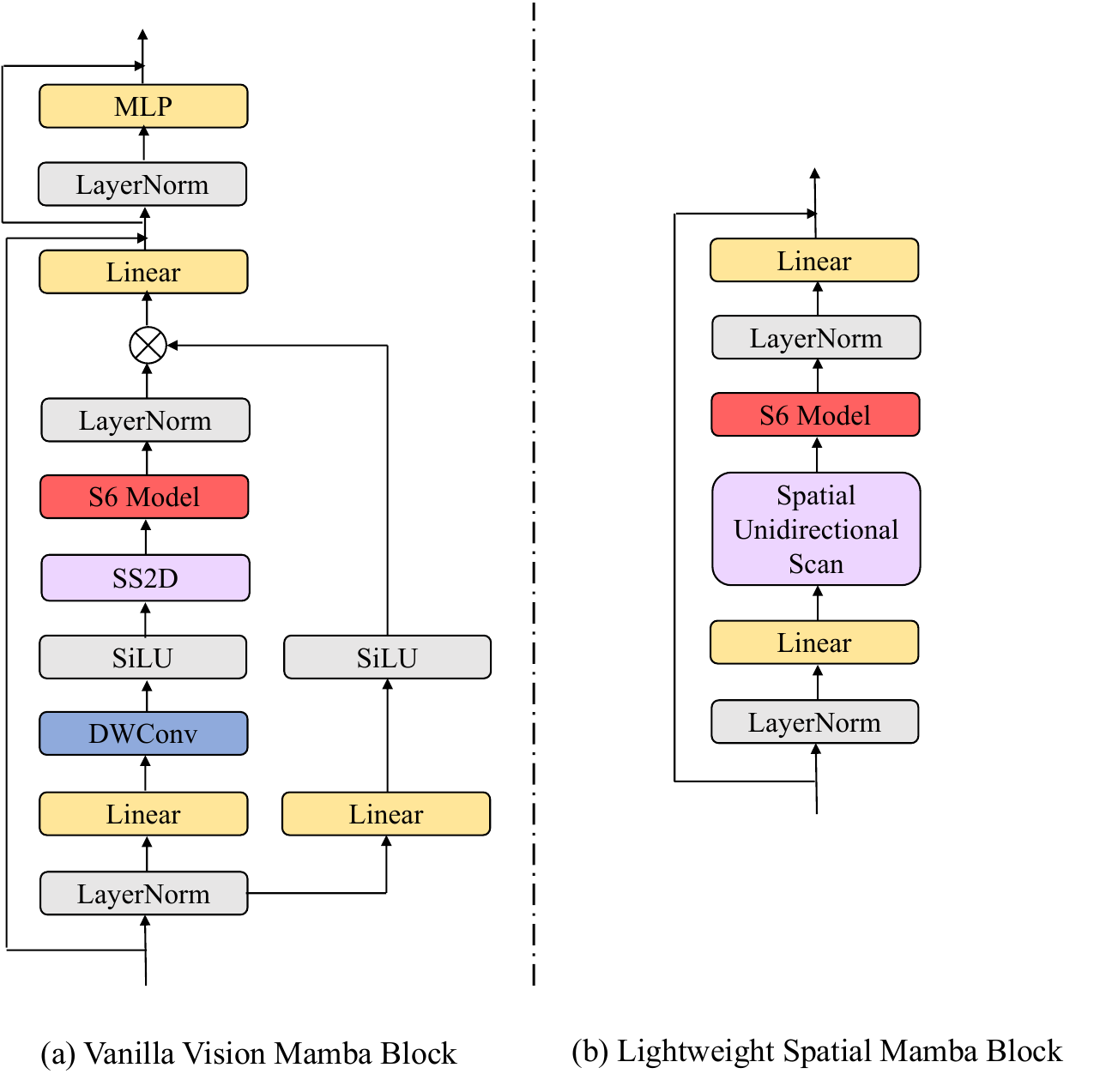}
    \caption{Illustration of the vanilla vision Mamba block in \cite{liu2024vmamba} and our proposed lightweight spatial Mamba block.}
    \label{fig3}
\end{figure}

\subsection{Cross-Attention Spectral-Spatial Mamba Module}
Mamba has demonstrated strong capabilities in global context modeling in NLP tasks and has shown initial success in 2D vision tasks. However, when dealing with HSI data, which exhibits complex spectral correlations and spatial information, existing vision Mamba methods struggle to effectively and efficiently extract and integrate both global spectral and spatial features to model spectral-spatial relations. Therefore, we propose the cross-attention spectral-spatial Mamba module. This module introduces dynamic positional embedding to provide spatial location information for the scanning sequence, and performs cross-attention spectral-spatial fusion, integrating spatial features and spectral features extracted by lightweight spatial/spectral Mamba block.
\subsubsection{Dynamic Positional Embedding} 
Since Mamba can only encode sequential data, it is necessary to transform 3D HSI data into sequence data through scanning. In this process, HSI data may lose positional information, yet modeling the global spatial context is location-aware. Therefore, it is imperative to introduce positional embeddings before scanning the HSI data to supplement spatial location information. Compared to absolute positional embeddings, dynamic positional embeddings are input-dependent and can dynamically adapt to the location information patterns of different types of inputs. Considering a lightweight design, we adopt depthwise convolution as our method for dynamic positional embedding (DPE):
\begin{equation}
DPE(X) = DWConv_{3 \times 3}(X),
\end{equation}
where $X \in \mathbb{R}^{P \times P \times D}$ is the pixel embedded feature, $P$ denotes patch size and $D$ denotes embedded dimension.

\subsubsection{Lightweight Spatial Mamba Block}
To efficiently capture global spatial contextual relations, we propose the lightweight spatial Mamba block. As illustrated in Fig.~\ref{fig3}, compared to the vanilla Vision Mamba block in \cite{liu2024vmamba}, we simplify the structure of the mamba block and eliminate the gated MLP, which relies on two linear transformations with a high parameter count to produce two features, thus slowing down computation. Additionally, we remove the large-parameter MLP and depthwise convolution, further enhancing computational efficiency. Instead, we first perform layer normalization on the pixel embedded feature $X$ with dynamic positional embedding $DPE(X)$, and then use only one linear layer for projecting the feature to the required dimensions of the S6 model:
\begin{equation}
\begin{aligned}
X_{pos} &= X + DPE(X),\\
X_{spa} &= Linear_1(LN(X_{pos})),
\end{aligned}
\end{equation}
where $X_{spa} \in \mathbb{R}^{P \times P \times D_{ssm}}$, $D_{ssm} = \text{ssm\_ratio} \times D$, $Linear_1(.) \in \mathbb{R}^{D \times D_{ssm}}$, $LN(.)$ denotes layer normalization.

For non-sequential vision data, existing vision Mamba methods typically scan the data horizontally and vertically across spatial dimensions to generate four spatial vision sequences, which are then processed by four separate S6 models to extract spatial features \cite{liu2024vmamba}. However, such a scanning strategy is inefficient for extracting spatial features from HSIs. Firstly, adding scans in each additional direction exponentially increases the number of parameters and FLOPs. Moreover, HSIs are acquired from high-altitude sensors with a nadir viewing angle, leading to similar spatial features irrespective of the scanning direction. Additionally, the scanned features are already enhanced with dynamic positional embeddings, ensuring that location information is incorporated at every spatial position. In light of this, we propose the spatial unidirectional scan strategy to efficiently extract global spatial features. As shown in Fig.~\ref{fig2}, we perform a pixel-by-pixel scan from left to right, generating a spatial vision sequence $S_{spa}=\{[x_0^{spa}, x_1^{spa}, ..., x_{P^2}^{spa}]|x_i^{spa} \in \mathbb{R}^{1 \times D_{ssm}}\}$. This sequence $S_{spa}$ is then fed into an S6 model for state space evolution as follows:
\begin{equation}
\begin{aligned}
h_t^{spa} &= \overline{\mathbf{A}}_{spa}h_{t-1}^{spa} + \overline{\mathbf{B}}_{spa}x_t^{spa}, \\
y_t^{spa} &= \mathbf{C}_{spa}h_t^{spa},
\label{eq:ssm_d_spa}
\end{aligned}
\end{equation}
where $\overline{\mathbf{A}}_{spa} \in \mathbb{R}^{D_{ssm} \times N}$, $\overline{\mathbf{B}}_{spa} \in \mathbb{R}^{B \times P^2 \times N}$, $\mathbf{C}_{spa} \in \mathbb{R}^{B \times P^2 \times N}$ are all the training parameters of the S6 model, and $N$ denotes the Mamba hidden state dimension.

After efficient spatial Mamba processing, we obtain an output sequence $Y_{spa}=\{[y_0^{spa}, y_1^{spa}, ..., y_{P^2}^{spa}]|y_i^{spa} \in \mathbb{R}^{1 \times D_{ssm}}\}$, and then perform a layer normalization and a linear projection to original embedded dimension $D$. Finally, the global spatial Mamba feature $G_{spa} \in \mathbb{R}^{P \times P \times D}$ is obtained with the residual connection from the original input as follows:
\begin{equation}
\begin{aligned}
G_{spa} = X_{pos} + Linear_2(LN(Y_{spa})), 
\label{eq:ssm_d_spa}
\end{aligned}
\end{equation}
where $Linear_2(.) \in \mathbb{R}^{D_{ssm} \times D}$.

\subsubsection{Lightweight Spectral Mamba Block}
Although the aforementioned block efficiently captures the global spatial context, it lacks consideration of the inherent continuous spectral band information in HSI data, which complicates the modeling of spectral-spatial relations. To efficiently capture the global spectral context, we propose the lightweight spectral Mamba block. The overall structure of this block is similar to the previously introduced lightweight spatial Mamba block, incorporating a thoughtful lightweight design. However, it differs in that it inputs the feature from the central pixel of an HSI patch to the block. This design choice is motivated by two main factors:
\begin{enumerate}
    \item The feature corresponding to the central pixel position directly reflects the spectral properties of the target surface feature and is less affected by spectral mixing.
    \item It significantly reduces the parameters and computational complexity of the model.
\end{enumerate}

Initially, we extract the center feature from the central pixel position of $X_{pos}$, and reshape it to the size of $(D, 1)$ Then, we perform a layer normalization on the center feature, and then project it to the dimensions required by the S6 model:
\begin{equation}
\begin{aligned}
X_{spe} &= Linear_3(LN(Center(X_{pos}))),
\end{aligned}
\end{equation}
where $X_{spe} \in \mathbb{R}^{D \times \text{ssm\_ratio}}$, $Linear_3(.) \in \mathbb{R}^{1 \times \text{ssm\_ratio}}$.

For spectral sequence features, given the asymmetry of spectral characteristics, capturing global spectral contextual information from both forward and reverse directions enables more comprehensive modeling of the interrelations and trends across different spectral bands. In view of this, we propose a spectral bidirectional scan strategy to efficiently extract global spectral features, as illustrated in Fig.~\ref{fig2}. We perform one scan in each direction along the spectral dimension, obtaining two spectral sequences $S_{spe}$ from the central spectral feature, $S_{spe}=\{[x_{0, j}^{spe}, x_{1, j}^{spe}, ..., x_{D, j}^{spe}]|x_{i, j}^{spe} \in \mathbb{R}^{1 \times \text{ssm\_ratio}}, j \in \{0, 1\}\}$. Sequences $S_{spe}$ are then fed into two S6 model for state space evolution as follows:
\begin{equation}
\begin{aligned}
h_{t, j}^{spe} &= \overline{\mathbf{A}}_{spe}^j h_{t-1, j}^{spe} + \overline{\mathbf{B}}_{spe}^j x_{t, j}^{spe}, \\
y_{t, j}^{spe} &= \mathbf{C}_{spe}^j h_{t, j}^{spe},
\label{eq:ssm_d_spe}
\end{aligned}
\end{equation}
where $\overline{\mathbf{A}}_{spe}^j \in \mathbb{R}^{\text{ssm\_ratio} \times N}$, $\overline{\mathbf{B}}_{spe}^j \in \mathbb{R}^{B \times D \times N}$, $\mathbf{C}_{spe}^j \in \mathbb{R}^{B \times D \times N}$ are all the training parameters of two S6 models, which are parameter-efficiency.

After efficient spectral Mamba processing, we obtain two output sequence $Y_{spe}=\{[y_{0, j}^{spe}, y_{1, j}^{spe}, ..., y_{D, j}^{spe}]|y_{i, j}^{spe} \in \mathbb{R}^{1 \times \text{ssm\_ratio}}, j \in \{0, 1\}\}$, and then merge them into one spectral sequence feature:
\begin{equation}
    Merge(Y_{spe}) = Y_{spe}^0 + Flip(Y_{spe}^1).
\end{equation}
We perform a layer normalization and a linear projection to the original embedded dimension. Finally, the global spectral Mamba feature $G_{spe} \in \mathbb{R}^{1 \times 1 \times D}$ is obtained with the residual connection from the original input as follows:
\begin{equation}
G_{spe} = X_{pos} + Linear_4(LN(Merge(Y_{spe}))), 
\label{eq:ssm_d_spe}
\end{equation}
where $Linear_4(.) \in \mathbb{R}^{\text{ssm\_ratio} \times 1}$.

\subsubsection{Cross-Attention Spectral-Spatial Fusion}
To effectively model the complex spectral-spatial relations, fully leveraging the complementary nature of spectral and spatial information to achieve a comprehensive spectral-spatial representation is essential. Therefore, we propose cross-attention spectral-spatial fusion to learn cross-correlation seamlessly integrating the global spectral/spatial Mamba feature. By leveraging a cross-attention mechanism, this approach enhances the complementarity between spectral and spatial features, guiding the model to effectively utilize existing features, reduce parameters, and improve computational efficiency.

Specifically, we normalize the global spectral Mamba feature $G_{spe}$ and the global spatial Mamba feature $G_{spa}$, respectively. This normalization process is intended to smooth the output of the softmax function, ensuring a fair calculation of the importance of different features. Subsequently, an average pooling operator is performed on $G_{spa}$ to obtain a comprehensive global spatial feature. We apply the softmax function to compute the spectral attention weight $A_{spe}$ and spatial attention weight $A_{spa}$:
\begin{equation}
\begin{aligned}
    A_{spe} &= Softmax(Norm(G_{spe})),\\
    A_{spa} &= Softmax(AP(Norm(G_{spa}))),
\end{aligned}
\end{equation}
where $AP(.)$ denotes the average pooling operation.
Then, the obtained spectral attention weight $A_{spe}$ and spatial attention weight $A_{spa}$ are cross-multiplied with the spatial feature $G_{spa}$ and spatial feature $G_{spe}$, respectively. Finally, the cross-attention features are summed to fuse into a comprehensive global spectral-spatial feature $G$:
\begin{equation}
\begin{aligned}
    G = A_{spe}G_{spa}+A_{spa}G_{spe} + X_{pos}
\end{aligned}
\end{equation}
\subsection{Lightweight Spectral-Spatial Residual Convolution Module}
Although Mamba efficiently captures global contextual information, modeling complex spectral-spatial relations necessitates a simultaneous consideration from both local and global perspectives. Therefore, we propose the lightweight spectral-spatial residual convolution module, which leverages the strong local feature extraction capabilities of CNNs to complement Mamba, enabling parallel and efficient modeling of spectral-spatial relations. Given the challenge of using a single lightweight convolution operator to extract both spectral and spatial features, we design two lightweight parallel branches, spectral branch and spatial branch.

Specifically, a lightweight 3D convolution is employed in the spectral branch to efficiently extract local spectral features. To facilitate the use of features in 3D convolution, an additional depth dimension is expanded on the features. We then utilize a 3D convolution operator with a kernel size of $1 \times 1 \times 3$, focusing the convolution primarily on the spectral dimension to aggregate information from adjacent spectral bands while preserving spatial information, thereby extracting local spectral features. Moreover, this convolution operator has only three parameters, making it extremely lightweight and low in FLOPs. The process can be formulated as follows:
\begin{equation}
    L_{spe} =  \delta(BN(Conv3d(X))),
\end{equation}
where $Conv3d(.)$ denotes the 3D convolution with a $1 \times 1 \times 3$ kernel, $BN(.)$ denotes batch normalization and $\delta(.)$ denotes SiLU activation function.

Another branch, the spatial branch, is designed with a $3 \times 3$ depthwise convolution to efficiently extract local spatial features. Depthwise convolution independently processes spatial features for each spectral channel without mixing spectral information, and with only $3 \times 3 \times D$ parameters, it facilitates efficient extraction of local spatial features. The process can be formulated as follows:
\begin{equation}
    L_{spa} =  \delta(BN(DWConv(X))),
\end{equation}
where $DWConv(.)$ denotes the depthwise convolution with a $3 \times 3$ kernel, $\delta(.)$ denotes SiLU activation function.

Subsequently, the extracted local spectral feature $L_{spe}$ is concatenated with the local spatial feature $L_{spa}$. Pointwise convolution is then used to fuse these features across channels, reducing the feature channels back to their original channels. The local spectral-spatial feature $L$ is obtained as follows:
\begin{equation}
    L =  \delta(BN(PWConv([L_{spe}, L_{spa}]))),
\end{equation}
where $PWConv(.)$ denotes the pointwise convolution, [.] denotes the concatenation and $\delta(.)$ denotes SiLU activation function.

\begin{figure}
    \centering
    \includegraphics[width = 0.9\columnwidth]{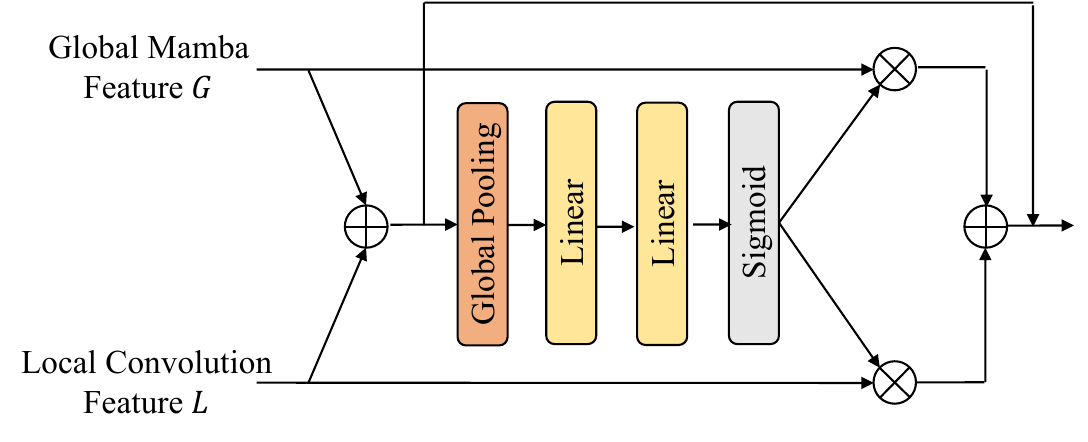}
    \caption{Structure of the proposed adaptive global-local fusion.}
    \label{fig4}
\end{figure}
\subsection{Adaptive Global-Local Fusion}
The adaptive fusion mechanism encourages the model to learn how to optimally integrate different types of features. This flexibility allows the model to not rely on any specific type of feature, enhancing its generalization capabilities across new land cover scenarios. To adaptively fuse the global spectral-spatial feature from the cross-attention spectral-spatial Mamba module and the local spectral-spatial feature from the lightweight spectral-spatial residual convolution module, we propose an adaptive global-local fusion module to obtain a comprehensive global-local spectral-spatial representation. The structure of this module is shown in Fig.~\ref{fig4}.

Specifically, We first fuse global spectral-spatial feature $G$ and local spectral-spatial feature $L$ via an element-wise summation and an average pooling operation, and then feed the fused feature into a simple multi-layer perception (MLP) to obtain a compact feature $z \in \mathcal{R}^{B \times 1}$ with fewer channels for better efficiency:
\begin{equation}
z = W_1(\delta(BN(W_2 \cdot AP(G+L)))),
\end{equation}
where $\delta(.)$ is the ReLU function, $BN(.)$ is the batch normalization, $W_1 \in \mathcal{R}^{K \times K_r}$, $W_2 \in \mathcal{R}^{K_r \times 1}$, $K_r = K/2$, $AP(.)$ denotes the average pooling operation.
The global feature weight $\mathcal{W}_g$ and the local feature weight $\mathcal{W}_l$ are calculated by the Sigmoid function as follows:
\begin{equation}
\begin{aligned}
    \mathcal{W}_g &= Sigmoid(z),\\
    \mathcal{W}_l &= 1 - \mathcal{W}_g.
\end{aligned} 
\end{equation}
To stabilize the training procedure, we add the extra skip connection from the prior fused feature. Finally, we obtain the comprehensive global-local spectral-spatial representation $F$ by multiplying the global/local weight:
\begin{equation}
\begin{aligned}
    F = G + L + \mathcal{W}_gG + \mathcal{W}_lL.
\end{aligned} 
\end{equation}

Finally, A linear classifier is employed to predict labels for HSI classification.
\begin{table}[t]
\centering
\caption{Land-cover types, the number of labeled training samples and testing samples of the Indian Pines dataset.}
\begin{tabular}{c||ccc}
\toprule[1.5pt]
Class&Land Cover Type&Training&Testing\\
\hline \hline  1&Alfalfa&5&41\\
 2&Corn-Notill&143&1285\\
 3&Corn-Mintill&83&747\\
 4&Corn&24&213\\
 5&Grass-Pasture&48&435\\
 6&Grass-Trees&73&657\\
 7&Grass-Pasture-Mowed&3&25\\
 8&Hay-Windrowed&48&430\\
 9&Oats&2&18\\
 10&Soybean-Notill&97&875\\
 11&Soybean-Mintill&245&2210\\
 12&Soybean-Clean&59&534\\
 13&Wheat&20&185\\
 14&Woods&126&1139\\
 15&Buildings-Grass-Trees-Drives&39&347\\
 16&Stone-Steel-Towers&9&84\\
\hline \hline &Total&1024&9225\\
\bottomrule[1.5pt]
\end{tabular}
\label{Table:Indian}
\end{table}

\begin{table}[t]
\centering
\caption{Land-cover types, the number of labeled training samples and testing samples of the WHU-Hi-Longkou dataset.}
\begin{tabular}{c||ccc}
\toprule[1.5pt]
Class&Land Cover Type&Training&Testing\\
\hline \hline 1     & Corn                & 172      & 34339   \\
2     & Cotton              & 42       & 8332    \\
3     & Sesame              & 15       & 3016    \\
4     & Broad-leaf soybean  & 316      & 62896   \\
5     & Narrow-leaf soybean & 21       & 4130    \\
6     & Rice                & 59       & 11795   \\
7     & Water               & 335      & 66721   \\
8     & Roads and houses    & 36       & 7088    \\
9     & Mixed weed          & 26       & 5203    \\
\hline \hline & Total               & 1022     & 203520 \\
\bottomrule[1.5pt]
\end{tabular}
\label{Table:longkou}
\end{table}

\begin{table}[t]
\centering
\caption{Land-cover types, the number of labeled training samples and testing samples of the Houston 2018 dataset.}
\begin{tabular}{c||ccc}
\toprule[1.5pt]
Class&Land Cover Type&Training&Testing\\
\hline \hline 1&Healthy Grass&490&9309\\
 2&Stressed Grass&1625&30877\\
 3&Artificial turf&34&650\\
 4&Evergreen trees&680&12915\\
 5&Deciduous trees&251&4770\\
 6&Bare earth&226&4290\\
 7&Water&13&253\\
 8&Residential buildings&1989&37783\\
 9&Non-residential buildings&11187&212565\\
 10&Roads&2293&43573\\
 11&Sidewalks&1702&32327\\
 12&Crosswalks&76&1442\\
 13&Major thoroughfares&2317&44031\\
 14&Highways&493&9372\\
 15&Railways&347&6590\\
 16&Paved parking lots&575&10925\\
 17&Unpaved parking lots&7&139\\
 18&Cars&327&6220\\
 19&Trains&269&5100\\
 20&Stadium seats&341&6483\\
\hline \hline &Total&25242&479614\\
\bottomrule[1.5pt]
\end{tabular}
\label{Table:H2018}
\end{table}

\section{Experiments and results}
In this section, we first describe three well-known public HSI datasets: the Indian Pines dataset, the Houston 2018 dataset, and the WHU-Hi-Longkou dataset. We then introduce the experimental setting, including evaluation metrics and implementation details. Following this, we undertake quantitative experiments and perform an ablation analysis to assess the efficacy of our proposed method.
\subsection{Datasets Description}
\subsubsection{Indian Pines}
The Indian Pines dataset, captured in 1992 by the Airborne Visible Infrared Imaging Spectrometer (AVIRIS) over an area of Indian pines in North-Western Indiana, comprises $145 \times 145$ pixels. Each pixel features a spatial resolution of 20 meters and spans 220 spectral bands within the wavelength range of 400 to 2500 nm. After excluding bands impacted by noise, 200 bands (1-103, 109-149, 164-219) are retained for classification purposes. This dataset includes 10,249 labeled pixels distributed across 16 categories. We use 10\% of the labeled samples for training and the rest for testing. The class name and the number of training and testing samples are listed in Table \ref{Table:Indian}.  

\subsubsection{WHU-Hi-Longkou}
The WHU-Hi-Longkou (Longkou) dataset, collected in 2018, utilizes an 8-mm focal length Headwall Nano-Hyperspec imaging sensor mounted on a DJ-Innovations Matrice 600 Pro UAV platform. This dataset encompasses $550 \times 400$ pixels, each with a spatial resolution of 0.463 meters, and includes 270 spectral bands ranging from 400 to 1000 nm in wavelength. It comprises 204,542 labeled samples across 9 object classes. We use 0.5\% of the labeled samples for training and the rest for testing. The class name and the number of training and testing samples are listed in Table \ref{Table:longkou}.

\subsubsection{Houston 2018}
Identified as the 2018 IEEE GRSS DFC dataset, the Houston 2018 dataset was collected by the National Center for Airborne Laser Mapping over the University of Houston campus and adjacent urban areas. The HSI component features $601 \times 2384$ pixels, each with a spatial resolution of 1 meter, covering 48 spectral bands in the wavelength range of 380 to 1050 nm. It encompasses 504,856 labeled pixels across 20 distinct classes of interest. We use 5\% of the labeled samples for training and the rest for testing. The class name and the number of training and testing samples are listed in Table \ref{Table:H2018}.

\subsection{Experimental Setting}
\subsubsection{Evaluation Metrics}
We evaluate the performance of all methods by three widely used indexes: overall accuracy (OA), average accuracy (AA), and Kappa coefficient ($\kappa$). 

\subsubsection{Comparison with State-of-the-art Methods}
To demonstrate the effectiveness of our proposed method, we compare our classification performance with several SOTA CNN-based, RNN-based, and transformer-based approaches using the most effective setting for these methods.
\begin{itemize}
\item The 2-D CNN \cite{yang2018cnn} architecture comprises three 2-D convolution blocks followed by a softmax layer. Each block includes a 2-D convolution layer, a batch normalization (BN) layer, an average pooling layer, and a ReLU activation function.
\item The 3-D CNN \cite{yang2018cnn} features three 3-D convolution blocks, each capped with a softmax layer. The blocks are constructed with a 3-D convolution layer, a BN layer, a ReLU activation function, and another 3-D convolution layer employing a stride of 2.
\item The SSRN \cite{zhong2018ssrn} is a spectral-spatial residual network that utilizes 3-D CNNs and residual connections. It includes both spectral and spatial residual blocks tailored to efficiently extract discriminative spectral-spatial features.
\item The AB-LSTM \cite{ablstm} is a bidirectional long short-term memory (Bi-LSTM)-based network utilizing a spatial-spectral attention mechanism.
\item The MSRT \cite{msrt} incorporates an RT encoder that merges RNNs (specifically, LSTMs) with the ViT into a unified architecture. It utilizes a multi-scanning-controlled positional embedding strategy to optimize feature fusion within the RT encoder.
\item In the SpectralFormer (SF) \cite{hong2021spectralformer}, group-wise spectral embedding and cross-layer adaptive fusion modules are integrated within the transformer framework. This approach is designed to capture local spectral representations from neighboring bands effectively.
\item The SSFTT \cite{sun2022ssftt} seamlessly combines CNN and transformer to leverage spectral-spatial information in HSI. It features a Gaussian-weighted feature tokenizer module, which enhances the separability of the samples.
\item The GAHT \cite{mei2022gaht} is a end-to-end group-aware transformer method with three-stage hierarchical framework.
\end{itemize}
\begin{table*}[ht]
\centering
\caption{Quantitative performance of different classification methods in terms of OA, AA, and $\kappa$ as well as the accuracies for each class on the Indian Pines dataset. The best results are shown in bold.}
\begin{tabular}{c||ccc|cc|ccc||c}
\toprule[1.5pt] \multirow{2}{*}{Class} & \multicolumn{3}{c|}{CNN-based Methods} & \multicolumn{2}{c|}{RNN-based Methods} & \multicolumn{3}{c||}{Transformer-based Methods} & \multirow{2}{*}{DualMamba} \\
\cline{2-9} & 2-D CNN & 3-D CNN & SSRN & AB-LSTM & MSRT & SF & SSFTT & GAHT \\
\hline \hline
1 & 65.85 & 58.54 & 94.14 & 37.07 & 96.81 & 63.00 & 95.12 & \bf97.56 & 96.83\\
2 & \bf99.77 & 76.19 & 97.84 & 85.82 & 98.48 & 92.35 & 97.67 & 98.05 & 99.07\\
3 & 81.66 & 77.64 & 97.54 & 82.72 & 94.38 & 86.86 & 98.87 & 98.66 & \bf99.00\\
4 & 96.71 & 52.11 & 90.70 & 65.77 & 95.22 & 88.96 & 91.55 & 95.31 & \bf98.69\\
5 & 85.75 & 93.56 & 97.75 & 88.14 & 97.66 & 92.49 & 96.32 & 95.17 & \bf97.91\\
6 & 97.87 & 98.17 & 99.24 & 97.14 & \bf100.00 & 99.12 & 99.54 & 99.85 & \bf99.85\\
7 & \bf100.00 & 36.00 & 81.60 & 32.40 & 95.29 & 52.50 & \bf100.00 & \bf100.00 & \bf100.00\\
8 & \bf100.00 & 98.60 & \bf100.00 & 96.49 & \bf100.00 & 99.16 & \bf100.00 & \bf100.00 & \bf100.00\\
9 & 50.00 & 55.56 & 74.44 & 12.78 & \bf100.00 & 41.18 & 88.89 & \bf100.00 & \bf99.44\\
10 & 35.54 & 82.86 & 94.77 & 82.22 & 95.88 & 93.16 & 97.71 & 94.29 & \bf98.29\\
11 & 88.01 & 90.45 & 98.87 & 90.36 & 96.89 & 92.27 & 98.69 & 99.37 & \bf99.58\\
12 & 98.13 & 62.55 & 97.83 & 73.60 & 97.95 & 85.44 & 98.13 & 96.63 & \bf99.01\\
13 & 99.46 & 88.65 & 99.24 & 94.86 & \bf100.00 & 99.02 & 97.28 & \bf100.00 & 99.73\\
14 & 99.91 & 99.39 & 99.18 & 97.32 & \bf100.00 & 96.73 & 99.91 & 97.89 & 99.72\\
15 & 91.35 & 86.17 & 93.95 & 68.65 & 92.55 & 83.41 & 98.84 & 97.12 & \bf98.96\\
16 & 86.90 & 45.24 & 98.33 & 94.17 & 98.51 & 93.50 & 95.54 & 94.05 & \bf99.64\\
\hline \hline
OA (\%) & 87.77 & 85.42 & 97.75 & 87.08 & 97.75 & 92.31 & 97.47 & 97.95 & \bf 99.23\\
AA (\%) & 86.06 & 75.10 & 94.71 & 74.97 & 97.48 & 84.95 & 96.57 & 97.75 & \bf 99.11\\
$\kappa$ & 0.8603 & 0.8324 & 0.9743 & 0.8526 & 0.9731 & 0.9124 & 0.9711 & 0.9766 & \bf 0.9912\\
\hline\hline
Params (K) & 488.84 & 1209.91 & 735.88 & 1577.51 & 1144.91 & 355.58 & 936.66 & 1038.16 & \bf72.94\\
FLOPs (M) & 24.78 & 213.98 & 210.71 & 14.42 & 77.41 & 61.39 & 79.94 & 53.63 & \bf4.19\\
\bottomrule[1.5pt]
\end{tabular}
\label{tab:IP}
\end{table*}

\begin{figure*} [thp]
	\centering
	\includegraphics[width=\textwidth]{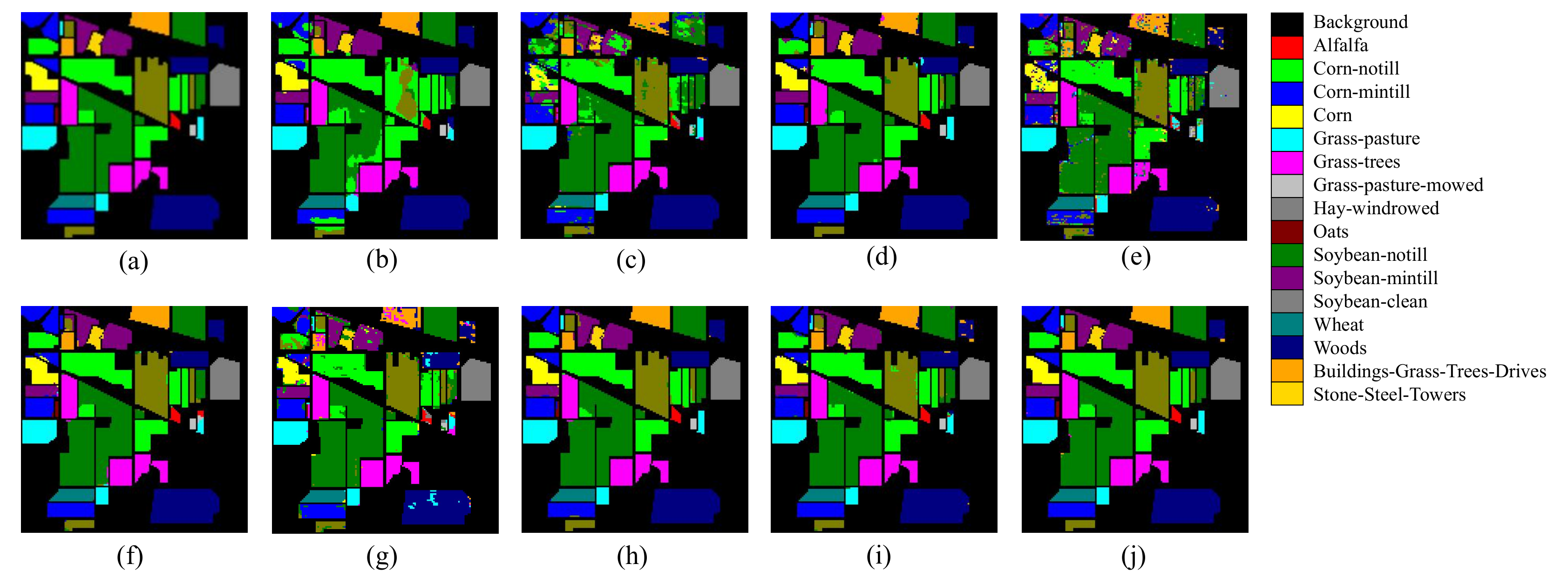}
	\caption{Classification maps obtained by different methods on the Indian Pines dataset. (b) 2-D CNN (OA=87.77\%). (c) 3-D CNN (OA=85.42\%). (d) SSRN (OA=97.75\%). (e) AB-LSTM (OA=87.08\%). (f) MSRT (OA=97.75\%). (g) SF (OA=92.31\%). (h) SSFTT (OA=97.47\%). (i) GAHT (OA=97.95\%). (j) DualMamba (OA=99.23\%).}
	\label{IP_results} 
\end{figure*}
\subsubsection{Implementation Details}
The proposed DualMamba is implemented with the Pytorch framework. The patch size is set to $7\times7$ for the Indian Pines dataset, $13\times13$ for the Longkou dataset and $15\times15$ for the Houston 2018 dataset. The AdamW optimizer is adopted with a batch size of 64 and a learning rate of 1e-3, training 300 epochs. We adopt the StepLR as our training schedule, in which the learning rate is multiplied by a gamma factor of 0.9 every 20 epochs. The embedded dimension $D$ is set to 64, Mamba hidden state $N$ is set to 16, and the $\text{ssm\_ratio}$ is set to 2. The initial pixel embedding is a $3 \times 3$ group convolution layer, in which the group number is 4. The cross-entropy loss is used for classification. We calculate the results fairly by averaging the results of ten repeated experiments with different training sample selections.

\begin{table*}[t]
\centering
\caption{Quantitative performance of different classification methods in terms of OA, AA, and $\kappa$, as well as the accuracies for each class on the WHU-Hi-Longkou dataset. the best results are shown in bold.}
\resizebox{0.85\textwidth}{!}{
\begin{tabular}{c||ccc|cc|ccc||c}
\toprule[1.5pt] \multirow{2}{*}{Class} & \multicolumn{3}{c|}{CNN-based Methods} & \multicolumn{2}{c|}{RNN-based Methods} & \multicolumn{3}{c||}{Transformer-based Methods} & \multirow{2}{*}{DualMamba} \\
\cline{2-9} & 2-D CNN & 3-D CNN & SSRN & AB-LSTM & MSRT & SF & SSFTT & GAHT \\
\hline \hline
1       & 99.44          & 99.55  & 99.80  & 99.13  & 99.33  & 99.76  & 99.85  & 99.91  & \textbf{99.96}  \\
2       & 95.44          & 93.18  & 98.82  & 89.04  & 99.41  & 89.20  & 95.96  & 98.99  & \textbf{99.66}  \\
3       & 81.76          & 93.83  & 91.11  & 89.48  & 97.94  & 97.25  & 93.50  & 96.68  & \bf98.42           \\
4       & \textbf{99.99} & 98.70  & 99.71  & 98.36  & 99.85  & 98.42  & 99.04  & 99.55  & 99.81           \\
5       & 75.45          & 83.74  & 94.04  & 82.98  & 91.26  & 83.80  & 92.42  & 95.96  & \textbf{98.01}  \\
6       & 98.66          & 98.88  & \bf99.89  & 96.46  & 99.81  & 97.69  & 99.34  & 99.77  & 99.86  \\
7       & 99.96          & \bf99.99  & 99.97  & 99.91  & 99.94  & 99.98  & \bf99.99  & \bf99.99  & 99.97  \\
8       & \textbf{98.41} & 96.66  & 96.33  & 92.57  & 91.93  & 89.45  & 97.19  & 96.46  & 96.78           \\
9       & 91.08          & 94.77  & 93.95  & 75.75  & 96.33  & 73.44  & 96.16  & 86.51  & \textbf{97.53}  \\
\hline \hline
OA (\%) & 98.58          & 98.48  & 99.28  & 97.28  & 99.20 & 97.47  & 99.02  & 99.19  & \textbf{99.66}  \\
AA (\%) & 93.36          & 95.37  & 97.07  & 91.52  & 97.31 & 92.11  & 97.05  & 97.09  & \textbf{98.89}  \\
$\kappa$ & 0.9812         & 0.9801 & 0.9905 & 0.9643  & 0.9895 & 0.9668 & 0.9872 & 0.9893 & \textbf{0.9956}\\
\hline \hline
Params (K) & 527.37 & 898.13 & 982.64 & 2761.63 & 1213.48 & 558.05 & 1258.69 & 1514.12 & \bf82.86\\
FLOPs (M) & 31.59 & 289.61 & 286.24 & 26.27 & 138.52 & 102.20 & 108.03 & 78.20 & \bf4.71\\
\bottomrule[1.5pt]
\end{tabular}}
\label{tab:Longkou}
\end{table*}

\begin{figure*} [thp]
	\centering
	\includegraphics[width=\textwidth]{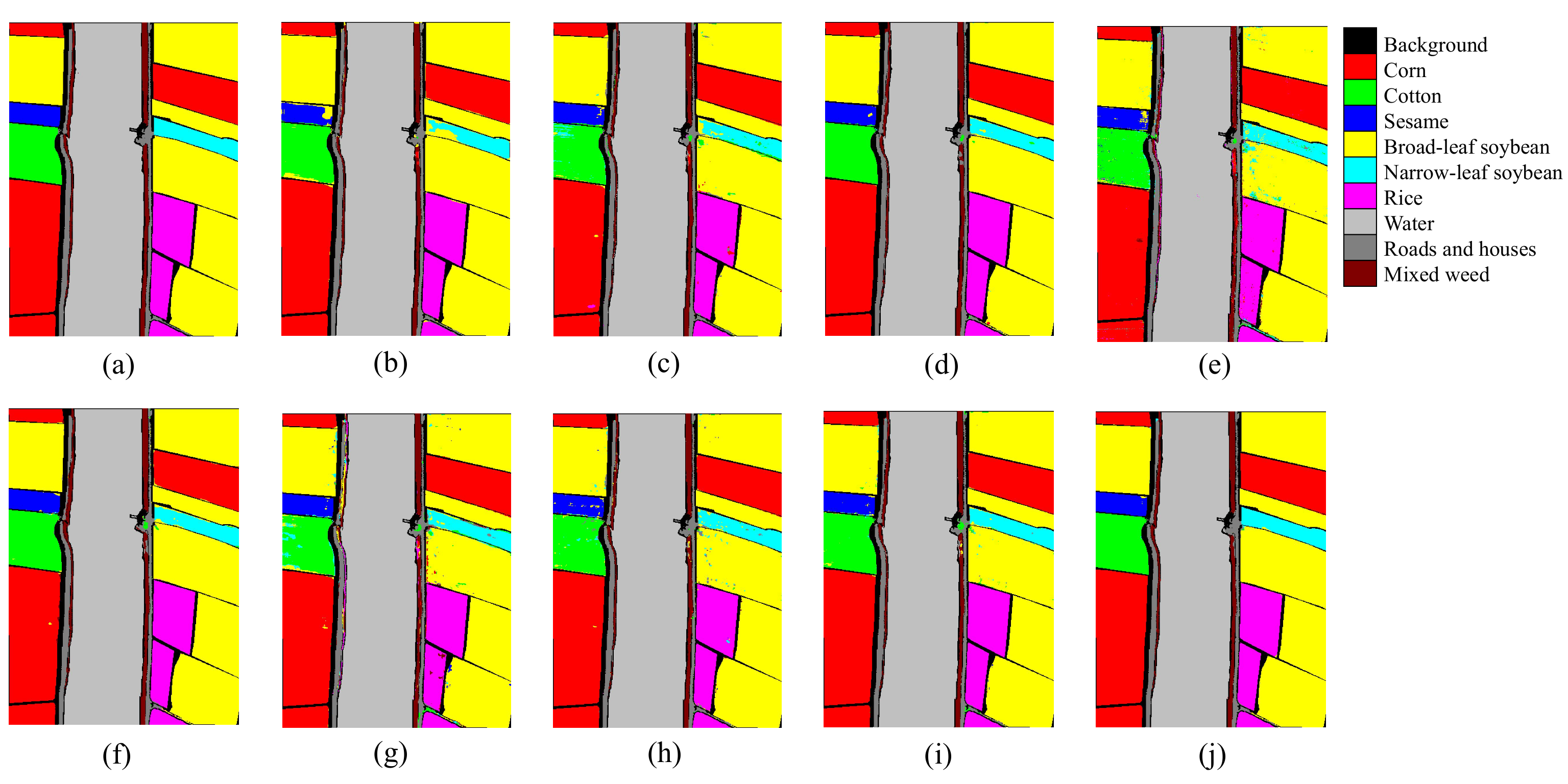}
	\caption{Classification maps obtained by different methods on the WHU-Hi-Longkou dataset. (a) Ground truth. (b) 2-D CNN (OA=98.58\%). (c) 3-D CNN (OA=98.48\%). (d) SSRN (OA=99.28\%). (e) AB-LSTM (OA=97.28\%). (f) MSRT (OA=99.20\%). (g) SF (OA=97.47\%). (h) SSFTT (OA=99.02\%). (i) GAHT (OA=99.19\%). (j) DualMamba (OA=99.66\%).}
	\label{LK_results} 

\end{figure*}

\begin{table*}[ht]
\centering
\caption{Quantitative performance of different classification methods in terms of OA, AA, and $\kappa$ as well as the accuracies for each class on the Houston 2018 dataset. the best results are shown in bold.}
\resizebox{0.85\textwidth}{!}{
\begin{tabular}{c||ccc|cc|ccc||c}
\toprule[1.5pt] \multirow{2}{*}{Class} & \multicolumn{3}{c|}{CNN-based Methods} & \multicolumn{2}{c|}{RNN-based Methods} & \multicolumn{3}{c||}{Transformer-based Methods} & \multirow{2}{*}{DualMamba} \\
\cline{2-9} & 2-D CNN & 3-D CNN & SSRN & AB-LSTM & MSRT & SF & SSFTT & GAHT \\
\hline \hline
1  & 87.12 & 82.02 & 86.30 & 90.32 & 88.44 & 92.36 & 79.93 & 79.50 & \bf93.57  \\
2  & 92.05 & 96.64 & 95.32 & 96.65 & 97.00 & 95.08 & 93.44 & 96.55 & \bf98.07  \\
3  & 96.92 & 96.00 & 99.72 & 98.92 & 99.85 & 96.21 & 99.66 & \bf100.00 & \bf100.00  \\
4  & 98.05 & 95.55 & 97.49 & 96.11 & 97.14 & \bf98.48 & 96.64 & 97.62 & 98.37  \\
5  & 87.25 & 79.16 & 86.09 & 81.93 & 93.98 & 91.87 & 90.11 & \bf95.01 & 92.52  \\
6  & 95.36 & 97.51 & 98.48 & 96.90 & 99.60 & 99.62 & 99.62 & 99.91 & \bf100.00 \\
7  & 96.44 & 71.94 & 93.68 & 77.08 & 98.12 & 27.01 & 85.45 & 95.65 & \bf98.81  \\
8  & 97.15 & 88.62 & 91.62 & 88.19 & 97.32 & 95.42 & 98.72 & 99.18 & \bf99.59  \\
9  & 98.24 & 92.80 & 97.88 & 96.61 & 98.71 & 98.60 & 99.09 & \bf99.35 & \bf99.35  \\
10 & \bf93.88 & 71.61 & 81.72 & 71.62 & 87.71 & 88.22 & 91.15 & 92.64 & 93.50  \\
11 & 75.85 & 73.17 & 69.44 & 67.66 & 79.17 & 27.01 & 80.97 & 85.73 & \bf88.74  \\
12 & 12.55 & 11.10 & 0.51  & 0.42 & 24.69 & 31.46 & 41.69 & 33.43 & \bf44.22  \\
13 & 85.24 & 69.12 & 84.59 & 85.63 & 92.53 & 91.79 & 94.49 & 96.79 & \bf97.46  \\
14 & 77.12 & 96.18 & 89.65 & 88.92 & 95.09 & 92.91 & 96.97 & \bf99.17 & 98.67  \\
15 & 94.45 & 98.98 & 99.34 & 99.32 & 99.62 & 99.33 & 99.30 & \bf99.92 & \bf99.92  \\
16 & 93.43 & 90.40 & 91.00 & 94.48 & 97.49 & 96.36 & 97.84 & 98.15 & \bf98.74  \\
17 & 64.75 & 20.86 & 0.00  & 46.04 & 99.92 & 22.78 & 69.21 & 90.65 & \bf100.00  \\
18 & 91.70 & 89.05 & 93.66 & 79.44 & 96.41 & 91.61 & 93.07 & \bf97.85 & 97.32  \\
19 & 96.88 & 95.45 & 96.92 & 91.90 & 98.41 & 96.53 & 97.97 & \bf99.84 & 99.61  \\
20 & 99.83 & 93.92 & 99.12 & 96.05 & 99.85 & 99.77 & 99.96 & \bf100.00 & 99.94  \\
\hline \hline
OA (\%) & 93.38 & 86.88 & 91.61 & 89.69 & 95.00 & 90.65 & 95.48 & 96.69 & \bf97.47 \\
AA (\%) & 86.71 & 80.50 & 82.63 & 82.21 & 92.14 & 80.75 & 90.26 & 92.85 & \bf94.92\\
$\kappa$ & 0.9137 & 0.8313 & 0.8906 & 0.8658 & 0.9349 & 0.8784 & 0.9412 & 0.9570 & \bf0.9670\\
\hline \hline
Params (K) & 402.31 & 427.54 & 199.74 & 219.60 & 1179.5 & 117.65 & 236.43 & 743.51 & \bf58.23\\
FLOPs (M) & 9.98 & 50.49 & 46.71 & \bf2.42 & 224.44 & 7.33 & 18.94 & 39.18 & 3.46\\
\bottomrule[1.5pt]
\end{tabular}}
\label{tab:HH}
\end{table*}

\begin{figure*} [thp]
	\centering
	\includegraphics[width=\textwidth]{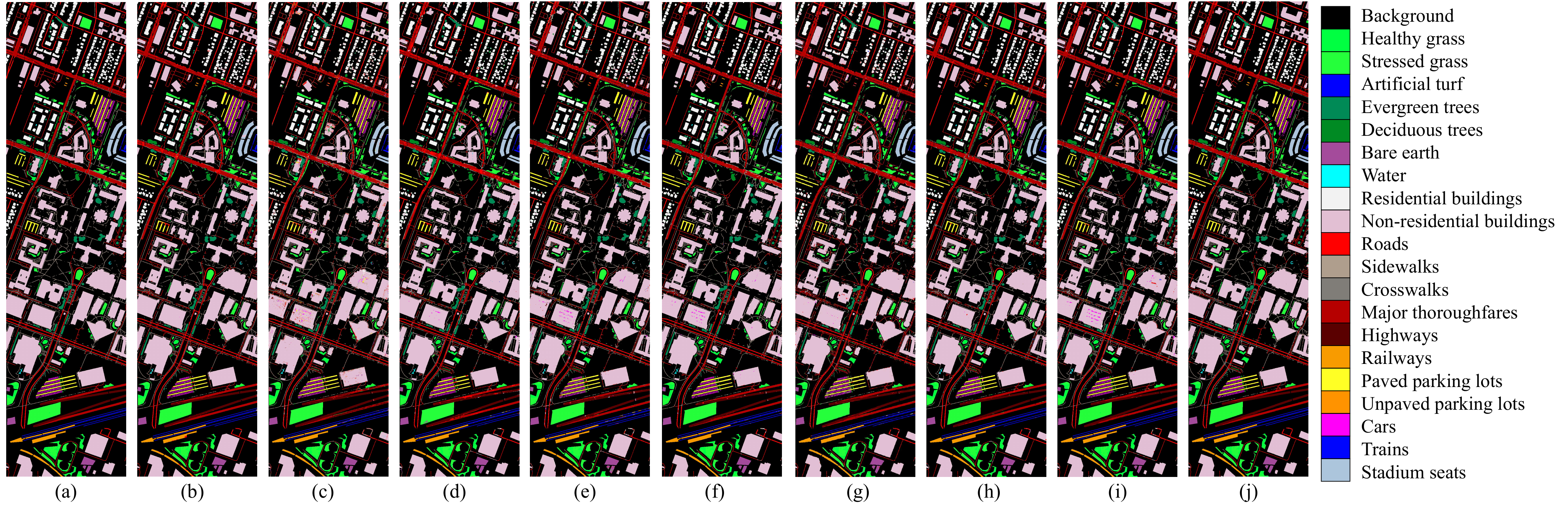}
	\caption{Classification maps obtained by different methods on the Houston 2018 dataset. (a) Ground truth. (b) 2-D CNN (OA=93.38\%). (c) 3-D CNN (OA=86.88\%). (d) SSRN (OA=91.61\%). (e) AB-LSTM (OA=89.69\%). (f) MSRT (OA=95.00\%). (g) SF (OA=90.65\%). (h) SSFTT (OA=95.48\%). (i) GAHT (OA=96.69\%). (j) DualMamba (OA=97.47\%).}
	\label{HU18_results} 

\end{figure*}
\subsection{Quantitative Results and Analysis}
\subsubsection{Classification Results Compared with SOTA Methods}
Quantitative results in terms of class-specific accuracy, overall accuracy (OA), average accuracy (AA), and Kappa and computational resources in terms of Parameters and FLOPs of different methods on the Indian Pines, Longkou, and Houston 2018 datasets are shown in Table \ref{tab:IP}–\ref{tab:HH}, respectively. The classification maps on the three datasets are shown in Fig. \ref{IP_results}–\ref{HU18_results}, respectively. 
Our proposed method achieves state-of-the-art performance across three datasets with minimal computational resource consumption, compared with other methods. Remarkably, our method outperforms the second-best approaches while using less than one-tenth of their parameters and FLOPs. Especially, on the Indian Pines dataset, our method achieves an overall accuracy of 99.28\%, compared to 97.95\% by the second-best method.
It can be observed that CNN-based methods, 2-D CNN, 3-D CNN, and SSRN, achieve good results due to their ability to capture local spectral-spatial information. However, their limited ability to extract global contextual features constrains their performance. RNN-based methods, AB-LSTM and MSRT are adept at modeling spectral relationships due to their proficiency in sequence problems, resulting in decent performance. Nevertheless, RNNs struggle with modeling long-term dependencies, which limits their effectiveness. Transformer-based methods, SF, SSFTT and GAHT, have good capabilities in representing global contextual features and achieving competitive results but also bring heavy computational overhead.
Our DualMamba, as a dual-stream hybrid Mamba-convolution network, exhibits good capability of capturing local spectral-spatial information as well as modeling long-range global contextual features through the cross-attention spectral-spatial Mamba module and lightweight spectral-spatial residual convolution module. Furthermore, the adaptive global-local fusion dynamically adjusts the weighting of global and local information, making it effective for HSI classification. 

\begin{figure*}[thp]
    \centering
    \subfloat[]{
		\includegraphics[width=0.32\textwidth]{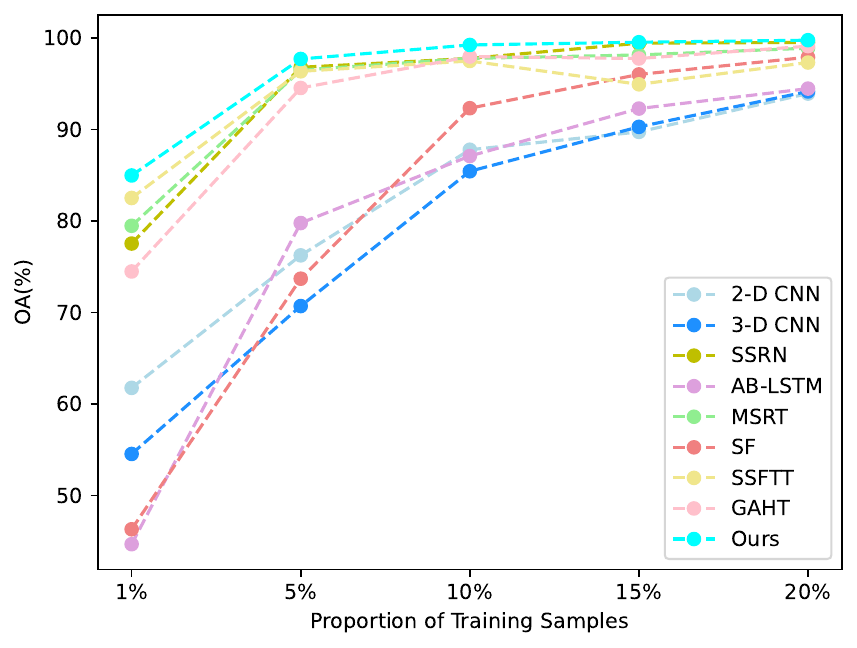}}
    \subfloat[]{
		\includegraphics[width=0.32\textwidth]{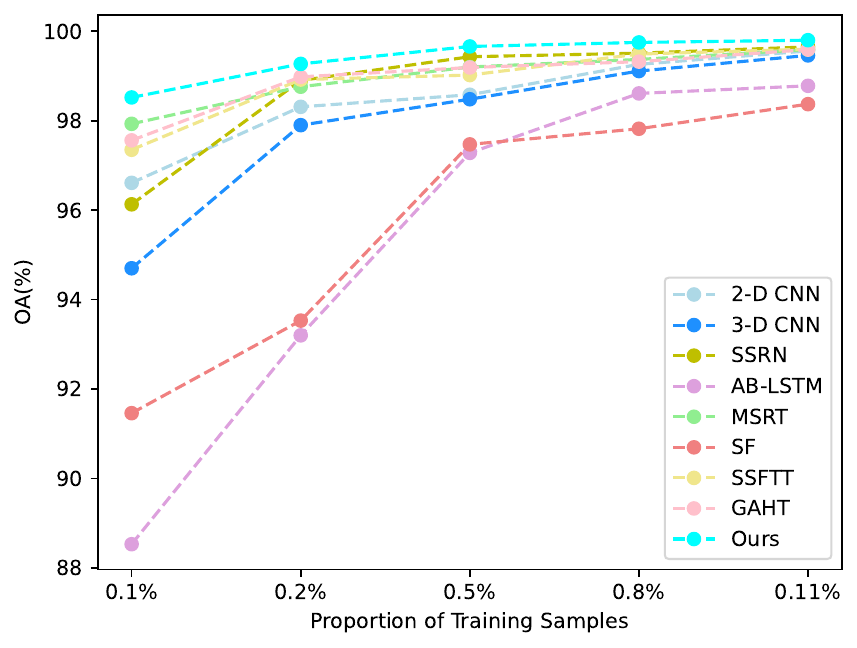}}
  \subfloat[]{
		\includegraphics[width=0.32\textwidth]{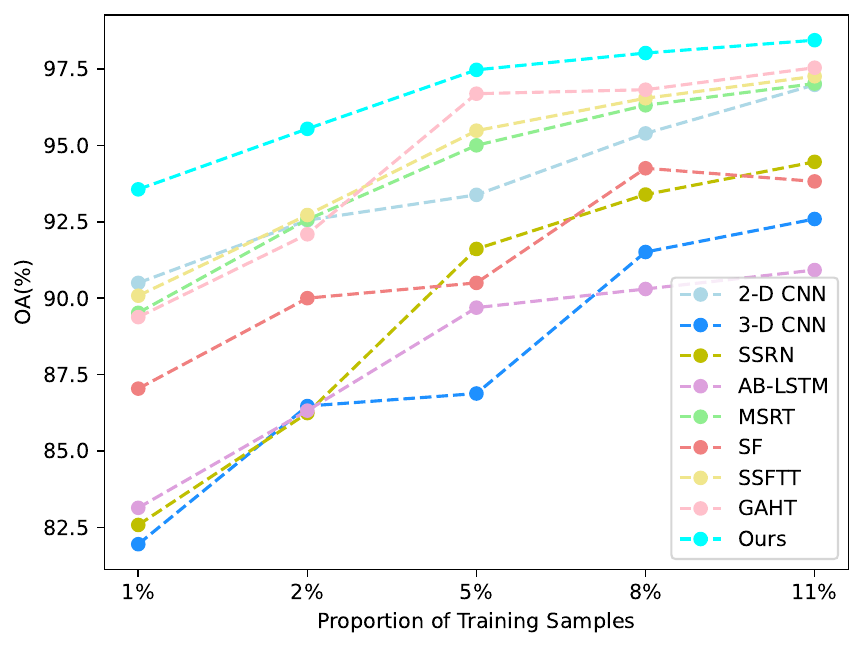}}
    \caption{Classification performance of the compared methods with different proportions of training samples on three datasets. (a) Indian Pines. (b) Longkou. (c) Houston 2018.}
    \label{fig:propotionresults}
\end{figure*}
\subsubsection{Classification Results with Different Proportions of Training Samples}
To further demonstrate the effectiveness of our method, extensive experiments by varying the proportions of training samples are conducted, with results shown in Fig.~\ref{fig:propotionresults}. It can be observed our method maintains a superior performance compared to other approaches that with different proportions of the training data in terms of OA on the three datasets. Especially when using only 0.5\% of the training samples, our model outperforms other leading methods that are trained with 0.11\% of the training samples in the Houston 2018 dataset. As the proportion of training samples increases, our method's performance steadily improves, showcasing its ability to effectively utilize additional data to enhance model performance.

\subsubsection{Model Complexity and Efficiency}
Table \ref{tab:IP}–\ref{tab:HH} display the parameters and FLOPs of the compared methods and our DualMamba, revealing that our proposed Mamba achieves the most lightweight design in the three datasets. Compared to CNN-based approaches, DualMamba achieves an average reduction of 88\% in parameters and 95\% in FLOPs. Compared to AB-LSTM, DualMamba exhibits a lesser reduction in FLOPs on the Houston 2018 dataset, due to the inherently lower computational overhead of RNNs when handling a smaller number of spectral channels. Nonetheless, our method still averages a 97\% decrease in parameters and 98\% in FLOPs compared to RNN-based methods. Serving as an alternative to transformers, Mamba models long-range dependencies with linear complexity, enabling DualMamba to attain a substantial decrease in parameters and FLOPs compared to transformer-based methods, averaging at 90\% fewer parameters and 98\% fewer FLOPs. These experimental outcomes affirm the success of our lightweight design, which concurrently delivers significant classification performance.
\subsection{Ablation Study and Analysis}
\begin{table*}[t]
\centering
\caption{Ablation for the Mamba-convolution Architecture across three datasets in terms of OA and parameters. The best results are shown in bold.}
\label{ab1}
\resizebox{0.8\textwidth}{!}{%
\begin{tabular}{cccccccc}
\hline
                        &                          & \multicolumn{2}{c}{Indian Pines}           & \multicolumn{2}{c}{Longkou}                 & \multicolumn{2}{c}{Houston 2018}            \\ \cline{3-8} 
\multirow{-2}{*}{{ CAS\textsuperscript{2}MM}} & \multirow{-2}{*}{{ LS\textsuperscript{2}RCMM}} & { OA (\%)} & { Params (K)} & { OA (\%)} & { Params (K)} & { OA (\%)} & { Params (K)} \\ \hline
{ \ding{51}}                        & { \ding{55}}                         & { 98.39}   & { 55.34}      & { 99.11}   & { 65.25}      & { 96.72}   & { 40.62}      \\
{ \ding{55}}                        & { \ding{51}}                         & { 97.06}   & { 43.41}      & { 98.37}   & { 53.33}      & { 92.45}   & { 28.69}      \\
{ \ding{51}}                        & { \ding{51}}                         & { \bf99.23}   & { 72.94}      & { \bf99.66}   & { 82.86}      & { \bf97.47}   & { 58.23}      \\ \hline
\end{tabular}%
}
\end{table*}

\begin{table*}[t]
\centering
\caption{Ablation for Components in Cross-Attention Spectral-Spatial Mamba Module across three datasets in terms of OA and parameters. The best results are shown in bold.}
\label{ab2}
\resizebox{0.97\textwidth}{!}{%
\begin{tabular}{cccccccccc}
\hline
\multirow{2}{*}{DPE} & \multirow{2}{*}{\begin{tabular}[c]{@{}c@{}}Lightweight Spatial \\ Mamba Block\end{tabular}} & \multirow{2}{*}{\begin{tabular}[c]{@{}c@{}}Lightweight Spectral \\ Mamba Block\end{tabular}} & \multirow{2}{*}{CAS\textsuperscript{2}F} & \multicolumn{2}{c}{Indian Pines} & \multicolumn{2}{c}{Longkou} & \multicolumn{2}{c}{Houston 2018} \\ \cline{5-10} 
                     &                                                                                             &                                                                                              &                        & OA (\%)       & Params (K)       & OA (\%)     & Params (K)    & OA (\%)       & Params (K)       \\ \hline
\ding{55}                    & \ding{51}                                                                                           & \ding{55}                                                                                            & \ding{55}                      & 98.45         & 72.09            & 99.12       & 81.87         & 96.62         & 57.24            \\
\ding{51}                    & \ding{51}                                                                                           & \ding{55}                                                                                            & \ding{55}                      & 98.64         & 72.73            & 99.23       & 82.51         & 96.97         & 57.88            \\
\ding{51}                    & \ding{51}                                                                                           & \ding{51}                                                                                            & \ding{55}                      & 99.10         & 72.94            & 99.50       & 82.86         & 97.18         & 58.23            \\
\ding{51}                    & \ding{51}                                                                                           & \ding{51}                                                                                            & \ding{51}                      & \bf99.23         & 72.94            & \bf99.66       & 82.86         & \bf97.47         & 58.23            \\ \hline
\end{tabular}%
}
\end{table*}
\subsubsection{Ablation for the Mamba-convolution Architecture}
The dual-stream Mamba-convolution architecture, central to DualMamba, has proven highly effective and efficient for modeling complex spectral-spatial relationships, as shown in Table \ref{ab1}. This architecture simultaneously captures global context through the cross-attention spectral-spatial Mamba module (CAS\textsuperscript{2}MM) and extracts local features using the lightweight spectral-spatial residual convolution Module (LS\textsuperscript{2}RCMM). Both the first and second experimental setups demonstrate that using either the Mamba branch alone or the CNN branch alone is inadequate for capturing the intricate features present in HSIs. Furthermore, the parallel Mamba-Convolution architecture does not severely increase the number of parameters, ensuring that the model remains lightweight and efficient.

\subsubsection{Ablation for Components in Cross-Attention Spectral-Spatial Mamba Module}
In the Cross-Attention Spectral-Spatial Mamba Module, four key components are critical: Dynamic Positional Embedding (DPE), Lightweight Spatial Mamba Block, Lightweight Spectral Mamba Block, and cross-attention spectral-spatial fusion (CAS\textsuperscript{2}F). Four sets of ablation experiments were designed and conducted across three datasets to evaluate the effectiveness of each component, with results displayed in Table \ref{ab2}. 

The first set of experiments, which solely utilizes the spatial Mamba block for extracting spatial features from HSIs, lacks spectral feature extraction, resulting in poorer performance. The second set is introduced with dynamic positional embedding, enhancing the spatial location information within the spatial sequences and improving performance. This is achieved with minimal parameter increase, just 0.64K, due to the lightweight design using depthwise convolution. The third set is utilized with lightweight Mamba blocks for both spectral and spatial dimensions to extract global contextual information and is performed with a simple addition for fusing spectral-spatial features. Extracting continuous spectral band information inherent to HSI data significantly enhances the ability to model spectral-spatial relations, thus greatly improving performance. The introduction of the lightweight spectral Mamba block adds only an average of 0.3K parameters, affirming our lightweight design's efficiency in extracting global spectral-spatial information. However, simple additive fusion was insufficient for modeling complex spectral-spatial relations. The fourth set of experiments demonstrated that our designed cross-attention spectral-spatial fusion, without introducing learnable parameters, can effectively leverage the complementarity of spectral and spatial information to achieve a comprehensive spectral-spatial representation, thus efficiently enhancing performance.

\begin{figure}[t]
    \centering
    \subfloat[Unidirectional]{
        \includegraphics[width=0.35\columnwidth]{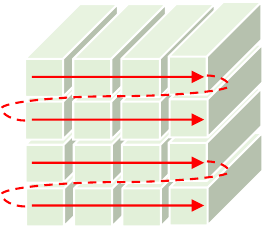}
        \label{fig:subfig1}
    }
    \quad\quad
    \subfloat[Bidirectional \cite{zhu2024visionmamba}]{
        \includegraphics[width=0.35\columnwidth]{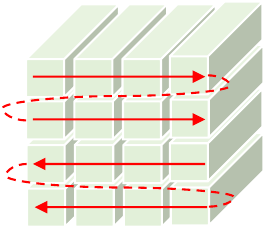}
        \label{fig:subfig2}
    }
    \\
    \subfloat[Four-way \cite{liu2024vmamba}]{
        \includegraphics[width=0.35\columnwidth]{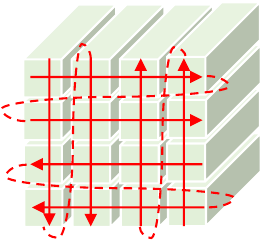}
        \label{fig:subfig3}
    }
    \quad\quad
    \subfloat[Omnidirectional \cite{chenhao2024omni}]{
        \includegraphics[width=0.35\columnwidth]{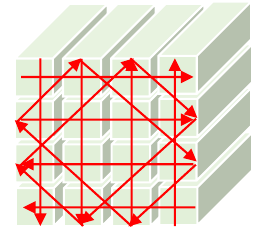}
        \label{fig:subfig4}
    }
    \caption{Illustration of existing spatial scanning strategies and our unidirectional spatial scan.}
    \label{fig:ab3}
\end{figure}

\begin{table}[t]
\centering
\caption{Ablation for Spatial Scan Strategy in terms of AA and parameters. The best results are shown in bold.}
\label{tab:scan}
\resizebox{\columnwidth}{!}{%
\begin{tabular}{ccccccc}
\hline
\multirow{2}{*}{\begin{tabular}[c]{@{}c@{}}Spatial Scan \\ Strategy\end{tabular}} & \multicolumn{2}{c}{Indian Pines} & \multicolumn{2}{c}{Longkou} & \multicolumn{2}{c}{Houston 2018} \\
                               & AA (\%)       & Params (K)       & AA (\%)     & Params (K)    & AA (\%)       & Params (K)       \\ \hline
Unidirectional                 & \bf99.11         & 72.94            & \bf98.89       & 82.86         & \bf94.92         & 58.23            \\
Bidirectional                  & 98.12         & 80.37            & 98.45       & 90.28         & 94.28         & 65.65            \\
Four-way                       & 98.87         & 95.22            & 98.41       & 105.13        & 94.45         & 80.50            \\
Omnidirectional                & 98.88         & 124.91           & 98.34       & 134.83        & 94.29         & 110.20           \\ \hline
\end{tabular}%
}
\end{table}

\begin{table}[t]
\centering
\caption{Ablation for Spectral Scan Strategy in terms of AA and parameters. The best results are shown in bold.}
\label{tab:spescan}
\resizebox{\columnwidth}{!}{%
\begin{tabular}{ccccccc}
\hline
\multirow{2}{*}{\begin{tabular}[c]{@{}c@{}}Spectral Scan \\ Strategy\end{tabular}} & \multicolumn{2}{c}{Indian Pines} & \multicolumn{2}{c}{Longkou} & \multicolumn{2}{c}{Houston 2018} \\
                               & AA (\%)       & Params (K)       & AA (\%)     & Params (K)    & AA (\%)       & Params (K)       \\ \hline
Unidirectional                 & 97.93         & 72.93            & 98.12       & 82.85         & 94.03         & 58.22            \\
Bidirectional                  & \bf99.11         & 72.94            & \bf98.89       & 82.86         & \bf94.92         & 58.23            \\
 \hline
\end{tabular}%
}
\end{table}
\begin{table}[t]
\centering
\caption{Ablation for Lightweight Mamba Block Design in terms of OA and parameters. The best results are shown in bold.}
\label{ab4}
\resizebox{\columnwidth}{!}{%
\begin{tabular}{ccccccc}
\hline
\multirow{2}{*}{}                                                    & \multicolumn{2}{c}{Indian Pines} & \multicolumn{2}{c}{Longkou} & \multicolumn{2}{c}{Houston 2018} \\
                                                                     & OA (\%)       & Params (K)       & OA (\%)     & Params (K)    & OA (\%)       & Params (K)       \\ \hline
\begin{tabular}[c]{@{}c@{}}Vanilla Vision\\ Mamba Block\end{tabular} & 99.11         & 115.65           & 99.55       & 125.57        & 97.19         & 100.94           \\
\begin{tabular}[c]{@{}c@{}}Lightweight\\ Mamba Block\end{tabular}    & \bf99.23         & 72.94            & \bf99.66       & 82.86         & \bf97.47         & 58.23            \\ \hline
\end{tabular}%
}
\end{table}
\subsubsection{Ablation for Scan Strategy}
To verify the effectiveness and efficiency of our unidirectional spatial scan in HSI classification, we compare it against three common scanning strategies used in existing vision Mamba methods: bidirectional scan, four-way directional scan, and omnidirectional scan, as shown in Fig.~\ref{fig:ab3}. These strategies incrementally add scanning sequences from different directions to enhance the capture of global spatial information. Although effective in other 2D vision tasks, for HSIs obtained from high-altitude sensors with a nadir viewing angle, the spatial features captured from any direction are similar, making these multi-directional scanning strategies redundant for HSI classification. Moreover, each additional scanning direction increases the use of the S6 model, thereby exponentially increasing the number of parameters and FLOPs, leading to computational inefficiency. Table~\ref{tab:scan} validates our analysis of the impact of scanning strategies on AA and the number of parameters. As the number of scanning directions increases, there is a large increase in parameters, and AA decreases compared to unidirectional scanning, which means overfitting. This demonstrates that our unidirectional spatial scanning strategy is both effective and efficient.

\begin{table}[t]
\centering
\caption{Ablation for Adaptive Global-Local Fusion across three datasets in terms of OA. The best results are shown in bold.}
\label{ab5}
\resizebox{\columnwidth}{!}{%
\begin{tabular}{cccc}
\hline
Fusion Strategy   & Indian Pines & Longkou & Houston 2018 \\ \hline
Sum                              & 98.87        & 99.41   & 96.93        \\
Concat + Linear                    & 98.94        & 99.45   & 97.02        \\
Learnable Weights                & 99.03        & 99.42   & 97.09        \\
Adaptive Fusion                  & \bf99.23        & \bf99.66   & \bf97.47        \\ \hline
\end{tabular}%
}
\end{table}
\begin{figure}[t!]
    \centering
    \includegraphics[width = 0.93\columnwidth]{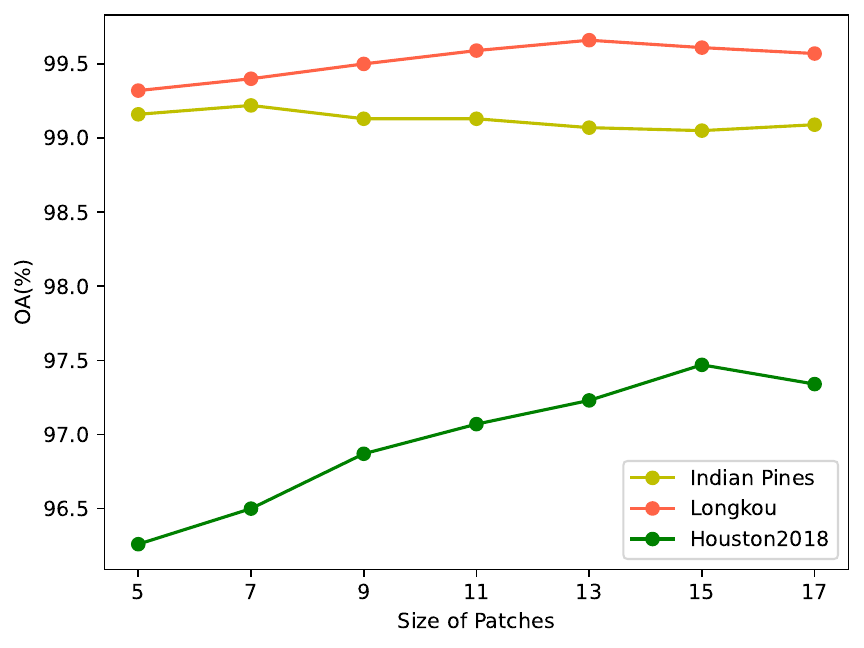}
    \caption{Classification results of different patch sizes on the Indian Pines, Longkou and Houston 2018 Dataset in terms of OA.}
    \label{fig:patch_size}
\end{figure}
Additionally, we conduct an ablation study for the spectral scanning strategy, as shown in Table \ref{tab:spescan}. Given the asymmetry of spectral characteristics, capturing global contextual information from both forward and reverse directions allows for more comprehensive modeling of the interrelations and trends across different spectral bands. Moreover, we limit our scanning to only the central features, thus additional scanning directions do not impose a heavy computational burden. Experimental results indicate that bidirectional scanning for spectral features outperforms unidirectional scanning, with only a minor increase of 0.1K in parameters, demonstrating the effectiveness and efficiency of spectral bidirectional scan.

\subsubsection{Ablation for Lightweight Mamba Block Design}
The ablation results for the lightweight Mamba block design across three datasets are shown in Table \ref{ab4}. We replaced the vanilla vision Mamba block \cite{liu2024vmamba} shown in Fig.~\ref{fig3} with our lightweight spectral/spatial Mamba block for experiments. The results indicate that our designed lightweight spectral/spatial Mamba block reduces the parameter count by an average of 42K across the three datasets, significantly decreasing computational consumption. Furthermore, there is a slight increase in OA, validating its effectiveness and parameter efficiency.

\subsubsection{Ablation for Adaptive Global-Local Fusion}
To validate the effectiveness of our proposed adaptive global-local fusion, we designed three fusion methods and conducted experiments on three datasets for comparison, as shown in Table \ref{ab5}. The first method involves a simple additive fusion of global spectral-spatial features with local spectral-spatial features. Such straightforward fusion can lead to an imbalance of information, as dominance of one type of feature may mask or neglect the other. The second method concatenates the two features and fuses them through a linear layer, allowing the model to learn correlations between the features but failing to effectively complement the global-local characteristics. The third method applies a learnable weight to each feature before their additive fusion. However, these learnable parameters are not flexible enough to accommodate different HSI patches and only yield relatively optimal weights for the relative dataset. In contrast, our method dynamically performs global-local fusion for each HSI sample, effectively integrating global contextual information from Mamba with local details from CNN.

\subsubsection{Parameter Analysis of Patch Size}
Evaluating our DualMamba across three datasets reveals distinct patch size dependencies, as shown in Fig. 9. For Indian Pines, OA peaks at 99.22\% with a patch size of 7, indicating a small patch size preference due to the smaller spatial size of this dataset. In contrast, Longkou's optimal performance occurs at a patch size of 13 with an OA of 99.66\%, suggesting a preference for moderately large inputs before a slight accuracy decline indicates potential overfitting. Houston 2018 shows a clear trend where increasing the patch size enhances performance, with the highest OA at 97.47\% for a patch size of 15. This trend suggests that larger patches capturing more contextual information are critical for accurately classifying complex urban landscapes.

\section{Conclusion}
In this paper, we propose DualMamba, an efficient and lightweight dual-stream hybrid Mamba-convolution network for hyperspectral image (HSI) classification. Our approach addresses the inherent challenges of high-dimensional spectral data and the need for computational efficiency in edge devices. By leveraging the global contextual modeling capabilities of the Mamba framework and the local feature extraction prowess of lightweight convolutional networks, DualMamba adeptly captures complex spectral-spatial relations from both global and local perspectives. Extensive evaluations on three public HSI datasets demonstrate that DualMamba significantly outperforms state-of-the-art methods, achieving superior classification accuracy with minimal parameters and FLOPs.


\bibliographystyle{IEEEtran}
\bibliography{ref} 

\vspace{11pt}




\vfill

\end{document}